\documentclass[journal]{IEEEtran}
\ifCLASSOPTIONcompsoc
\usepackage[nocompress]{cite}
\else
\usepackage{cite}
\fi
\usepackage{multirow}
\usepackage{xcolor,colortbl}
\definecolor{mygreen}{RGB}{0,255,0}
\definecolor{mygrey}{RGB}{192,192,192}

\ifCLASSINFOpdf
\usepackage[pdftex]{graphicx}
\else
\fi
\usepackage{amsmath}
\usepackage{array}
\usepackage{siunitx}
\usepackage{dcolumn}
\usepackage{float}
\usepackage{comment}
\usepackage{amsmath,mathrsfs}
\usepackage{amsfonts}
\usepackage{hyperref}
\usepackage{bbding}
\usepackage{caption}
\usepackage{subfigure}

\newcommand{\mytablespace}{1.15}

\hypersetup{
	colorlinks=true,
	linkcolor=blue,
	filecolor=blue,      
	urlcolor=blue,
	citecolor=mygreen,
}

\newcolumntype{d}[1]{D{.}{.}{#1}}

\begin{document}

\title{CARNet: Compression Artifact Reduction for Point Cloud Attribute}
\author{Dandan Ding, Junzhe~Zhang, Jianqiang~Wang, and Zhan Ma
        \thanks{D.~Ding and J.~Zhang are with the School of Information Science and Technology, Hangzhou Normal University, Hangzhou, Zhejiang 311121, China.}
		\thanks{J.~Wang and Z.~Ma is with the School of Electronic Science and Engineering, Nanjing University, Nanjing, Jiangsu, China.}
		\thanks{This research was supported by Zhejiang Provincial Natural Science Foundation of China under Grant LY20F010013 and National Natural Science Foundation of China under Grant 62171174.}
		}

\maketitle

\begin{abstract}
%The abstract goes here.
A learning-based adaptive loop filter is developed for the Geometry-based Point Cloud Compression (G-PCC) standard to reduce attribute compression artifacts.
The proposed method first generates multiple Most-Probable Sample Offsets (MPSOs) as potential compression distortion approximations, and then linearly weights them for artifact mitigation. As such, we drive the filtered reconstruction as close to the uncompressed PCA as possible. To this end, we devise a Compression Artifact Reduction Network (CARNet) which consists of two consecutive processing phases: MPSOs derivation and MPSOs combination. The MPSOs derivation uses a two-stream network to  model local neighborhood variations from  direct spatial embedding and frequency-dependent embedding, where sparse convolutions are utilized to best aggregate information from sparsely and irregularly distributed points. The MPSOs combination is guided by the least square error metric to derive weighting coefficients on the fly to further capture content dynamics of input PCAs. The CARNet is implemented as an in-loop filtering tool of the G-PCC, where those linear weighting coefficients are encapsulated into the bitstream with negligible bit rate overhead. Experimental results demonstrate significant improvement over the latest G-PCC both subjectively and objectively. For example, our method offers 21.96\% YUV BD-Rate (Bj{\o}ntegaard Delta Rate) reduction against the G-PCC across various commonly-used test point clouds. Compared with a recent work showing state-of-the-art performance, our work not only gains 12.95\% YUV BD-Rate but also provides 30$\times$ processing speedup.  
\end{abstract}


\IEEEpeerreviewmaketitle

\section{Introduction} 
\label{sect:introduction}
\IEEEPARstart{P}{oint} cloud is a collection of excessive number of points that are sparsely and irregularly distributed in the 3D space. It can flexibly and realistically represent a variety of 3D objects and scenes in many applications, such as Augmented Reality/Virtual Reality (AR/VR) and  autonomous driving~\cite{haala2008mobile,schwarz2018emerging}. Every valid point in a point cloud has its geometric coordinate, e.g., ($x,y,z$) in Cartesian coordinate system, and associated attribute component, such as the RGB color, normal, and reflectance. This work primarily deals with the color attribute. Unlike 2D images where pixels are uniformly-distributed and well structured for the characterization of spatial correlations easily, points in a 3D point cloud are unstructured, which makes it practically difficult to learn local neighborhood correlations for compact representation. This, in return, hinders the networked point cloud applications.

\begin{table}[t]
\renewcommand{\arraystretch}{1.15} 
    \centering
        \caption{Notations}
    \label{tab:notations}
    \begin{tabular}{c|c}
    \hline
       Abbreviation  &  Description \\
    \hline
     BD-Rate & Bj{\o}ntegaard Delta Rate~\cite{BDrate}\\
     CAR    & Compression Artifact Reduction\\
     G-PCC  & Geometry-based PCC\\
     LSE & Least Square Error\\
     MPEG & Moving Picture Expert Group\\
     MPSO & Most-Probable Sample Offset\\
     PCA      & Point Cloud Attribute\\
     PCAC  & Point Cloud Attribute Compression\\
     PCC      & Point Cloud Compression\\     
     \hline
    \end{tabular}

\end{table}

To this end, since 2017, the Moving Picture Experts Group (MPEG) of International Standards Organization (ISO) has been intensively investigating and promoting potential technologies for high-efficiency point cloud compression (PCC), leading to the conclusion of two PCC specifications, a.k.a.,  Geometry-based PCC (G-PCC) and Video-based PCC (V-PCC)~\cite{cao2021compression,cao20193d,graziosi_nakagami_kuma_zaghetto_suzuki_tabatabai_2020}. In V-PCC, a 3D point cloud sampled at a specific time instance is first projected to a set of perpendicular 2D planes; then a sequence of 2D planes consecutively spanning over a period of time are compressed using standard compliant video codecs like the prevalent HEVC (High-Efficiency Video Coding)~\cite{sullivan2012overview} or the latest VVC (Versatile Video Coding)~\cite{VVC_overview}. On the other hand, G-PCC directly compresses the 3D point cloud by separating its geometry and attribute components: the well-known octree model is often used for representing geometry coordinates, and Region-Adaptive Hierarchical Transform (RAHT)~\cite{de2016compression} and Hierarchical Prediction as Lifting Transform (PredLift) are selective options to compress the attribute information lossily~\cite{graziosi_nakagami_kuma_zaghetto_suzuki_tabatabai_2020}.

\subsection{Background and Motivation}
For the lossy compression of PCC using either V-PCC or G-PCC, compression artifacts like quantization-induced blockiness and blurriness are inevitable, which is annoying for perceptual appearance. Since the projection-based V-PCC adopts the HEVC or VVC as the compression backbone, rules-based in-loop filters~\cite{VVC_Inloop} including deblocking, sample adaptive offset (SAO), and/or adaptive loop filter (ALF) adopted in video codecs already significantly mitigate compression artifacts. Besides, recently-emerged learning-based filters can be augmented on top of existing codecs as either a post-processing module~\cite{ma2020mfrnet,guan2019mfqe} or an in-loop function~\cite{ding2021neural, nasiri2021cnn} to further improve the quality of restored pixels (e.g., YUV or RGB). 
{\it This work thus focuses on the compression artifact reduction (CAR) of G-PCC coded point cloud attribute (PCA)}. {Similar to existing works~\cite{zhang2014point,sheng2022attribute,schwarz2018emerging}, we assume the geometry information has been losslessly reconstructed when dealing with the lossy attribute compression.}

\begin{figure}[t]
  \centering
  \subfigure[]{{\includegraphics[width=0.27\linewidth]{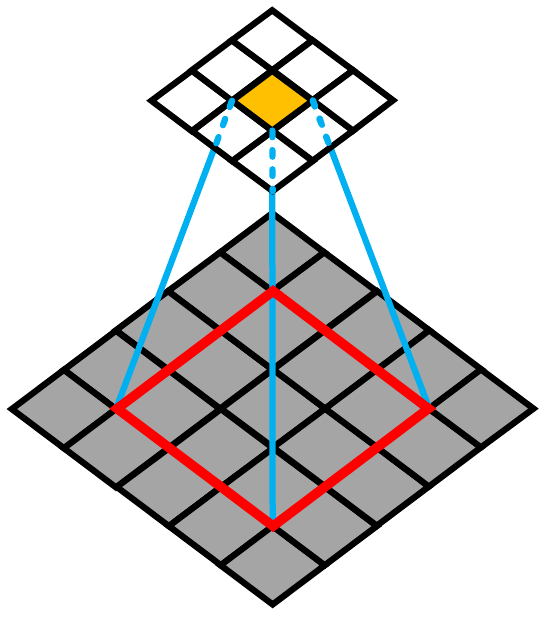}}}
  \subfigure[]{{\includegraphics[width=0.71\linewidth]{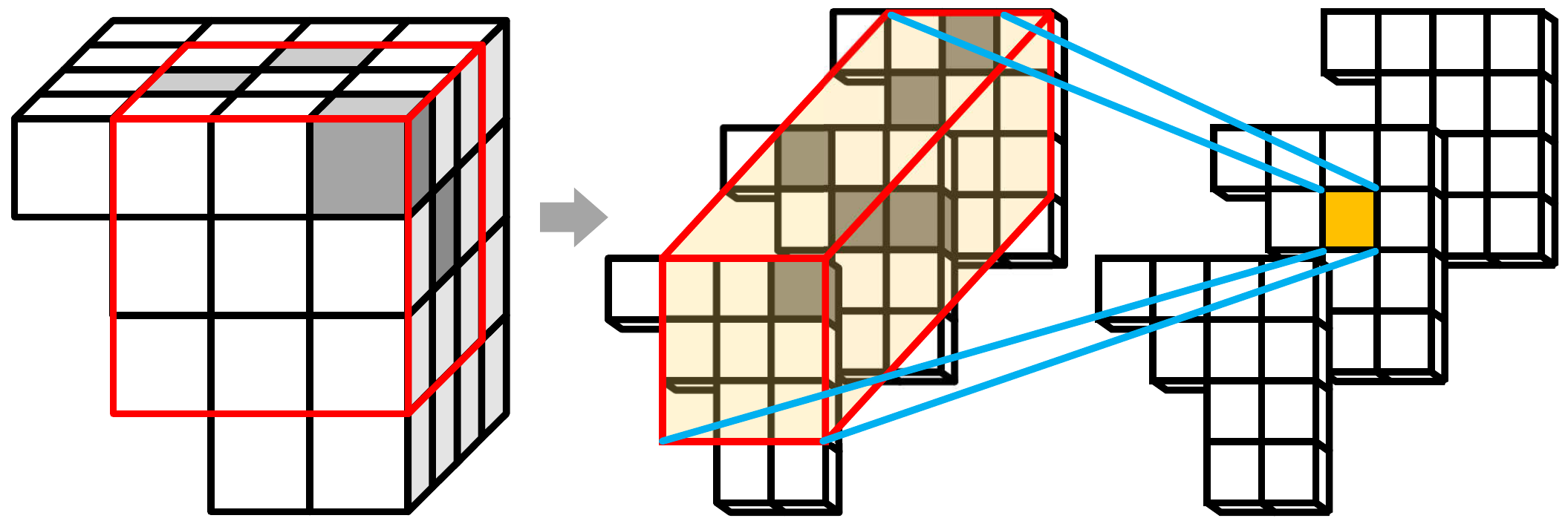}}}
  \caption{{\bf Characterization of Spatial Variations Across Local Neighbors.} (a) Neighbors in a $3\times3$ square window of a pixel in a 2D image. (b) Potential neighbors in a $3\times3\times3$ cube of a positively-occupied voxel (or point) in a 3D point cloud. Grey cubes stand for positively-occupied voxels which are converted from the raw points by voxelizing the input point cloud~\cite{schwarz2018emerging}. Whether a voxel is occupied is nondeterministic and highly content-dependent because of the dynamic, sparse, and unstructured distribution of points in a point cloud.}
  \label{fig:2D_3D_convolution}
\end{figure}

Unfortunately, either rules-based or learning-based filters were seldom applied to reduce the attribute artifacts of G-PCC compressed point clouds. This is mainly because it is hard to effectively characterize and model attribute relations across sparsely and irregularly distributed points in a local neighborhood.  As a naive comparison shown in Fig.~\ref{fig:2D_3D_convolution}, in an image,  $n\times n$\footnote{We assume the use of square window for neighborhood characteristic modeling. Other window shapes like diamond can be used as well~\cite{VVC_Inloop,tsai2013adaptive}.} neighbors centered at a specific pixel can be used to model local neighborhood variations to develop a 2D ALF for artifact removal; similarly, for a 3D point cloud, $n\times n \times n$ neighbors centered at a specific point (or positively-occupied voxels\footnote{Since raw point clouds are voxelized for the compression using G-PCC, the term ``point'' in a raw point cloud is equivalent to the ``positively-occupied voxel'' in its voxelized version. Therefore this work often uses these two terms interchangeably.}) can be potentially utilized to develop a proper 3D filter. On one hand, as $n$ enlarges, the complexity demanded for computing and caching $n^3$ elements increases much faster than that of processing $n^2$ elements, even for a simple convolutional operation. And more importantly, due to the sparse, unorganized, and irregular distribution nature, the occupancy status of neighboring voxels in a $n\times n\times n$ cube is nondeterministic and dynamic (see Fig.~\ref{fig:2D_3D_convolution}), which further complicates the characterization of attribute variations from spatial neighbors. Although devising an extremely large-scale network might be capable of learning such spatial variations, the complexity is accordingly unbearable for practical applications~\cite{sheng2022attribute}. 

This work, therefore, develops an efficient attribute artifact reduction filter to overcome difficulties aforementioned and improve the quality of compressed PCA with low complexity consumption.

\subsection{Approach}
Thanks to the superior representation capability of deep learning technologies, we resort to Deep Neural Networks (DNNs) for the implementation of a learned  3D in-loop filter to restore the quality of compressed PCAs. It is referred to as the CARNet (Compression Artifact Reduction Network).

%{\it Local Neighborhood Modeling.}
In principle, spatial coherency shall be maintained across neighborhood pixels in a 2D image or the attributes of neighboring points in a 3D point cloud even after compression, with which a visually pleasant appearance is ensured for content rendering~\cite{tsai2013adaptive}. In this regard, we propose to leverage spatial characteristics learned from a local neighborhood to develop the CARNet. Similar to those in-loop filtering studies for compressed 2D image/video~\cite{VVC_Inloop,tsai2013adaptive,SAO}, our CARNet also compensates the attribute distortion by estimating additive sample offsets, targeting to approach the original PCA input $\mathbf{{F}}$ as close as possible after the filtering process, i.e.,
\begin{align}
   \arg \min\nolimits_\Theta || (\mathbf{{F}} - \hat{\bf F}) - \mathcal{G}_{\Theta}(\hat{\bf F})||^2, \label{eq:carnet_alf}
\end{align} where $\hat{\bf {F}}$ is the compressed PCA, and $\mathcal{G}_{\Theta}(\cdot)$ stands for the proposed CARNet adapted by a set of parameters $\Theta$. As seen, ${\bf e} = \mathbf{{F}} - \hat{\bf F}$ is compression-induced PCA distortion.

The derivation of $\mathcal{G}_{\Theta}(\cdot)$ is proceeded as follows:
\begin{itemize}
    \item For each point (or positively-occupied voxel) $p$, we define the 3D local neighborhood centered at it as $\Phi_p$. Since point clouds are voxelized for G-PCC compression, we use a cubic $n\times n\times n$ to identify the range of  $\Phi_p$ in this work for simplicity. Here we only use positively-occupied voxels inside $\Phi_p$ to learn local characteristics for $p$. With this aim, we utilize sparse convolutions and stack them to build up sparse DNNs to aggregate valid local neighbors in $\Phi_p$.  To simplify the implementation, the range of $\Phi_p$ can be set the same as the receptive field of underlying sparse convolutions.

    \item To thoroughly model the spatial coherency in a local neighborhood, we devise a two-stream network in the CARNet, where one stream applies sparse DNNs to directly characterize local variations spatially, and the other stream first separates high- and low-frequency components to embed frequency-dependent neighborhood relations and then concatenates them together for subsequent feature fusion. 

    \item To best estimate the distortion $\bf e$ of a compressed PCA, we propose to generate multiple most-probable sample offsets (MPSOs) by aggregating two-stream features. These MPSOs are then linearly combined through the Least Square Error (LSE) optimization as in Eq.~\eqref{eq:carnet_alf} to best approach the compression distortion. Note that the linear weighting coefficients are derived on-the-fly and encapsulated into bitstream with negligible bit rate consumption for each frame of input point clouds.

\end{itemize}

\subsection{Contribution}

The main contributions of this paper are summarized below:

\begin{itemize}
    \item An efficient and low-complexity CARNet is developed to reduce the artifacts of G-PCC compressed point cloud attributes.  Extensive results demonstrate that the CARNet brings 21.96\% BD-Rate reduction to the latest G-PCC anchor across various common test point clouds recommended by the MPEG standardization committee. Relative to state-of-the-art MS-GAT~\cite{sheng2022attribute} which is a post-processing solution for artifact removal, our CARNet provides 12.95\% BD-Rate gains and costs much less runtime (about 30$\times$ speedup). 
    
    \item The efficiency of the CARNet comes from the effective characterization of neighborhood variations for the derivation of the most-probable sample offsets, which are then linearly weighted to produce the best additive offset for compression distortion compensation. The linear weighting coefficients are calculated on-the-fly through the guidance of original PCA. As such, we can best capture the dynamics of underlying content for better model generalization.

    \item The lightweight computation of the CARNet is owing to the use of sparse convolutions to aggregate positively-occupied neighbors only within the convolutional receptive field. This not only best leverages the sparseness of valid points in a point cloud but also significantly increases the computational efficiency and reduces the time complexity, as has also been extensively studied for point cloud geometry processing~\cite{PCGCv2,SparsePCGCv1,thanh2022learning}.
\end{itemize}

\section{Related Work} 
\label{sect:related_work}
This section reviews relevant studies on point cloud attribute compression and compression artifact reduction. 

\subsection{Point Cloud Attributes Compression}
\label{subsec:attribute_compression}

The transform coding framework has been widely used for the compression of point cloud attribute~\cite{952802,gu20193d}. For example, Zhang {\it et al.} applied Graph Transform (GT)~\cite{zhang2014point} to compress color attributes on small graphs constructed by nearby points. However, GT is computationally expensive because it requires eigenvalue decomposition. Later, Ricardo {\it et al.}~\cite{de2017transform} proposed Gaussian Process Transforms (GPTs), which are equivalent to Karhunen-Lo{\'e}ve Transforms (KLTs) of the Gaussian Process. More or less the same time, another Region-Adaptive Hierarchical Transform (RAHT)~\cite{de2016compression}, which is basically an adaptive Haar wavelet transform, was developed.
Note that RAHT was adopted into the MPEG G-PCC as the main tool for lossy attribute compression. Recently, the prediction of RAHT coefficients~\cite{souto2020predictive} was studied and included in the latest G-PCC reference software TMC13v14 with state-of-the-art compression efficiency reported. Besides, the hierarchical neighborhood Prediction as Lifting transform, termed PredLift, was also suggested in the G-PCC as the other lossy configuration of attribute coding.

Built upon recent advances in deep learning techniques, DNNs can also be used to facilitate the transform coding. Instead of applying handcrafted rules, data-driven learning is applied to derive (non-linear) transforms and context models directly for PCA compression. Among them, end-to-end supervised learning is the most straightforward solution~\cite{quach2022survey}. 
Sheng~{\it et al.}~\cite{sheng2021deep} designed a point-based lossy attribute autoencoder, where stacked multi-layer perceptrons (MLPs) were used to extract spatial correlations across points and transform the input attribute into high-dimensional features for entropy coding. He~{\it et al.}~\cite{he2022density} introduced a density-preserving deep point cloud compression framework, which could be extended to support attribute compression.  Wang~{\it et al.}~\cite{PCAC} proposed an end-to-end sparse convolution-based PCA compression method, called SparsePCAC, for high-efficiency feature extraction and aggregation. Currently, although these end-to-end learning-based solutions demonstrate encouraging potentials, their compression performance is still inferior to the latest G-PCC reference model TMC13v14.

In addition to the end-to-end approach, Fang~{\it et al.}~\cite{fang20223dac} used an MLP-based model to replace the traditional entropy coding tool in G-PCC. They encoded  RAHT coefficients using a neural model where side information including tree depth, weight, location, etc., was leveraged to estimate the probability of each coefficient for arithmetic coding. 

The above rules-based and learning-based solutions all involve the quantization operation for attribute compression. This inevitably introduces annoying compression artifacts and leads to unsatisfied perceptual sensation, imposing the urgent requirement for artifact reduction and quality improvement of compressed PCAs.

\subsection{Compression Artifact Reduction Methods}

Compression artifact reduction was extensively studied for restoring better reconstruction quality of compressed 2D images/videos, including both rules-based and learning-based methods, for either in-loop filtering or post-processing~\cite{HEVC_dblk,SAO,tsai2013adaptive,VVC_Inloop,nasiri2021cnn,jpeg_dong2015compression, ma2020mfrnet, wang2021multi, lin2022nr, pan2020efficient, jia2021residual, zhang2018residual, wang2021combining, kong2020guided}. Recently, learning-based solutions presented outstanding performance with remarkable quality improvement, which is mainly due to the use of powerful DNNs that effectively model spatial or spatiotemporal neighborhood characteristics for quality enhancement and artifact removal~\cite{ding2021neural,guan2019mfqe}.

In spite of abundant learning-based solutions for the quality enhancement of compressed 2D images/videos, we cannot intuitively extend them to support the quality enhancement of 3D PCAs. Unlike well-structured pixels in 2D images/videos, points in a pint cloud are sparsely and irregularly distributed. As a result, it is much more difficult to exploit the local neighborhood variations in 3D point clouds than in 2D images/videos. To tackle this problem, the MS-GAT~\cite{sheng2021deep}, a pioneering and probably the only published exploration on the artifacts removal of G-PCC compressed PCAs, adopted graph convolution and graph attention layers instead of trivial MLPs for attribute correlation exploration. As reported in MS-GAT, using a 1.98 MB post-processing network achieved 10.76\%, 6.14\%, and 8.83\% BD-Rate gains over the G-PCC for Y, U, and V components, respectively, at the expense of extremely high computational complexity. Even slicing the input point cloud to block patches with each patch having only 2048 points, its running speed is about 380$\times$ slower than the G-PCC anchor\footnote{Because the G-PCC and MS-GAT adopt different implementation platform and techniques, the runtime results only serve as the intuitive reference to have a general idea about the computational complexity. }, according to the complexity report of MS-GAT. In this regard, an efficient-yet-lightweight compression artifact reduction approach is highly desired.

\begin{figure*}[t]
  \centering
  \centerline{\includegraphics[width=\linewidth]{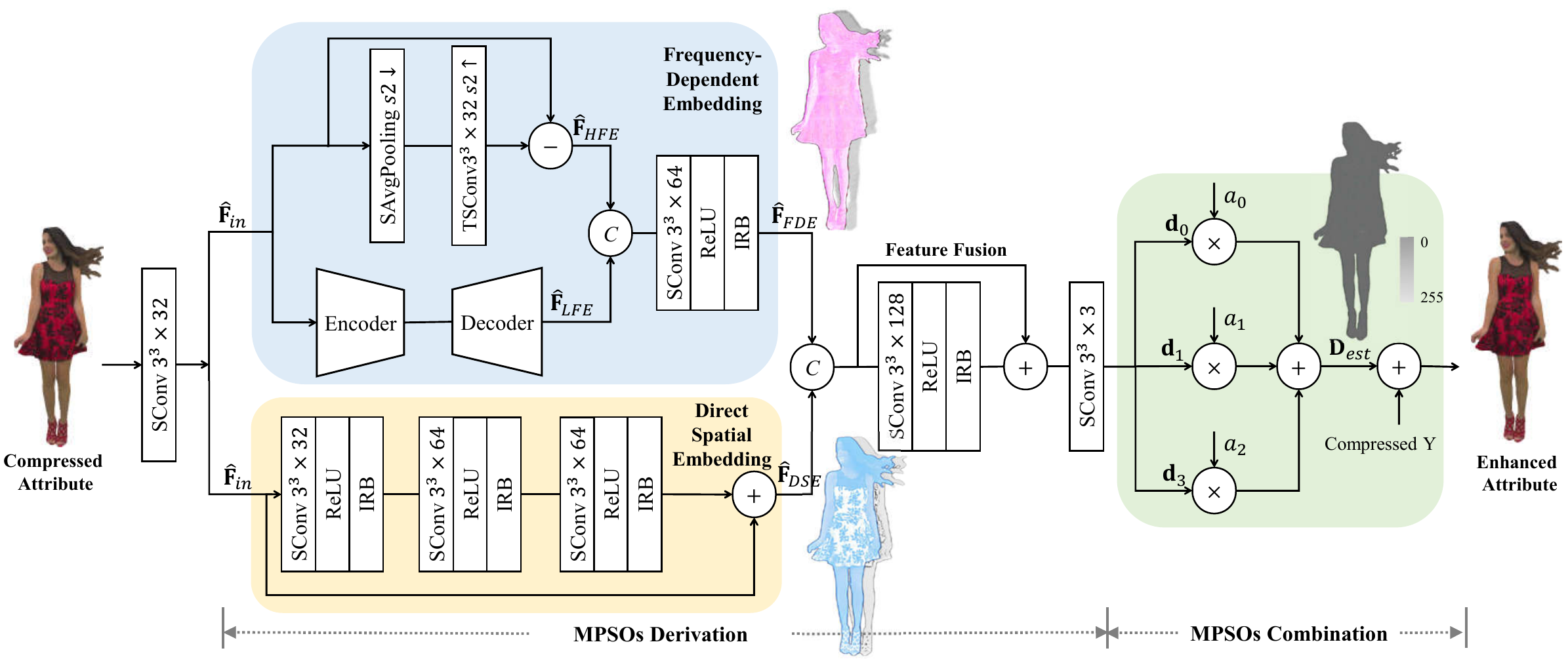}}
  \caption{{\bf The proposed {Compression Artifact Reduction Network~(CARNet)}.} The CARNet consists of two processing phases: the MPSOs derivation and the MPSOs Combination. The term ``MPSOs'' stands for Most-Probable Sample Offsets. The first phase uses a two-stream network to derive multiple MPSOs as potential approximations of attribute compression distortion; subsequently, the second phase leverages least square error optimization to linearly combine  MPSOs for attribute artifact compensation and quality improvement. The proposed two-stream network for the MPSOs derivation has two branches: one is for Frequency-Dependent Embedding and the other is for Direct Spatial Embedding, and the MPSOs are produced by fusing features from these two branches.  
  ``C'', ``+'', and ``-'' indicate concatenate, sum, and subtract operations, respectively. SConv $k^3\times C$ is the spatial convolution with $k\times k\times k$  kernel size  and $C$ channels. Simple ReLU is used for activation. The IRB (Inception Residual Block) stacks sparse convolutions with different depths for multi-level feature aggregation. 
  }
  \label{fig:CARNet_framework}
\end{figure*}

\section{The Proposed CARNet for Artifact Reduction of Compressed Point Cloud Attributes}
\label{sect:proposed_method}

Without loss of generality,  a given point cloud is represented using a sparse tensor ${\bf S} = ({\bf C}, {\bf F})$, where ${\bf C} = \{\vec{C}_i$ = ($x_i$, $y_i$, $z_i$), $i\in[0, N-1]\}$ stands for the collection of geometry coordinates, and ${\bf F} = \{\vec{F}$ = ($Y_i$, $U_i$, $V_i$), $i\in[0, N-1]\}$ represents associated color attributes. Following the convention used in existing works~\cite{zhang2014point,graziosi_nakagami_kuma_zaghetto_suzuki_tabatabai_2020},  point cloud geometry coordinates are losslessly compressed  as the prior knowledge to construct local neighborhood for the lossy compression of color attributes.

\subsection{Framework of the CARNet}
\label{subsect:framework}
As aforementioned, we propose the CARNet to solve Eq.~\eqref{eq:carnet_alf}. The CARNet is comprised of two consecutive phases: MPSOs Derivation and MPSOs Combination, as illustrated in Fig.~\ref{fig:CARNet_framework}. The MPSOs Derivation phase uses a two-stream network to derive a group of Most-Probable Sample Offsets as potential approximations of attribute compression distortion; and the MPSOs Combination phase linearly weights MPSOs to produce the best additive sample offset for distortion compensation of compressed PCAs.

Note that the derivation of MPSOs fully relies on the sparse DNNs to characterize and embed spatial variations in a local neighborhood under the spatial coherent assumption on attributes. The linear weighting coefficients of MPSOs are computed on-the-fly, further generalizing the CARNet to instantaneous PCA input.

First, the compressed three-channel PCA in YUV format, i.e., $\hat{\bf F}\in \mathcal{R}^{N\times 3}$, is projected to a high-dimensional feature space as $\hat{\bf F}_{in}\in \mathcal{R}^{N\times C}$ for subsequent processing using sparse convolution ``SConv $k^3\times C$'' with $k\times k \times k$ kernel and $C$ channels. $N$ is the total number of valid points (or positively-occupied voxels) in this compressed PCA. 

Since the significant difference between Y and U/V components, this work separately processes Y, U and V\footnote{The separate processing of Y, U, and V is also applied in MS-GAT~\cite{sheng2022attribute}.}. When processing the U component, it is concatenated with the compressed Y component for filtering; similarly, both compressed Y and U components are concatenated with V for its enhancement. In this way, we basically leverage the cross-component processing for better performance~\cite{Cross-Component}. As seen in Fig.~\ref{fig:preprocessUV}, the same CARNet architecture is used but model parameters are adapted accordingly through the training of different color components. Next, we detail the CARNet for the Y component first, and then extend it to the processing of U/V components.

\subsection{MPSOs Derivation}
\label{subsect:Sparse_Convolution_Based_Feature_Aggregation}

\begin{figure}[t]
  \centering
  \centerline{\includegraphics[width=0.57\linewidth]{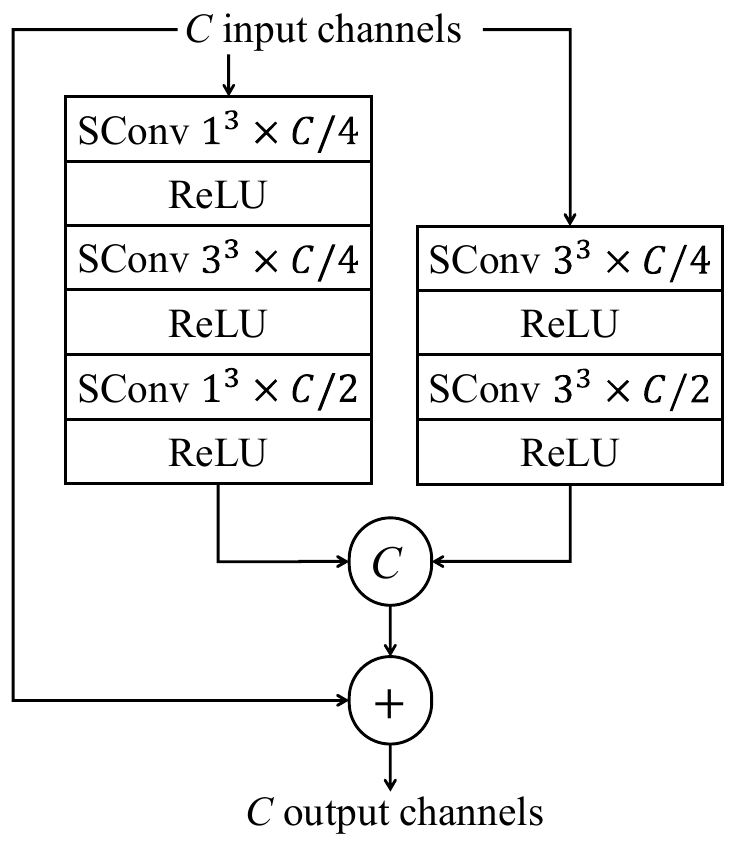}}
  \caption{{\bf Inception Residual Block (IRB)} used for efficient multi-level feature aggregation. Sparse convolution (SConv $k\times C$ is stacked with $k\times k$ kernel and $C$ channels. Simple ReLU is used for activation. }
  \label{fig:IRB}
\end{figure}

A Two-Stream network is developed to accept $\hat{\bf F}_{in}$ separately into two branches  for the generation of a group of MPSOs to approximate the compression distortion.

{\bf Direct Spatial Embedding (DSE).} 
One branch of the proposed two-stream network directly aggregates useful information from spatial neighbors in close proximity to model spatial variations. This generally assumes that the color attributes are more or less similar to  each other across spatial neighbors nearby. To best characterize and embed spatial correlations of PCAs, we cascade three identical Direct Spatial Embedding units to generate high-dimensional feature tensors.  Each each unit is comprised of a convolutional layer using a $3\times 3$ kernel, an activation layer using simple ReLU, and an Inception Residual Block (IRB). The IRB module is borrowed from our previous work~\cite{PCGCv2} for its efficiency to aggregate information from multi-level representations. A typical example of the IRB is shown in Fig.~\ref{fig:IRB}. As seen, we stack sparse convolutional layers with different depth hierarchies, kernel sizes, and channel numbers for the aforementioned multi-level information embedding. Moreover, both the IRB and the Direct Spatial Embedding branch are augmented by residual link~\cite{he2016deep} for fast and robust aggregation of output features $\hat{\bf F}_{DSE}$.

{\bf Frequency-Dependent Embedding (FDE).} In addition to directly analyzing and embedding spatial characteristics,  we separate the input $\hat{\bf F}_{in}$ to perform frequency-dependent embedding. This is because that compression-induced distortion usually presents different levels of artifacts across various frequency bands. For example, visually-annoying artifacts often reside in regions with  high-frequency components of the input signal, such as object edges and contours, since low-frequency components can be well represented by transforms and/or predictions, while high-frequency components easily suffer from imperfect prediction and  degradation in quantization. In this regard, we propose individual Low-Frequency Embedding and High-Frequency Embedding.

\begin{figure}[t]
  \centering
  \centerline{\includegraphics[width=\linewidth]{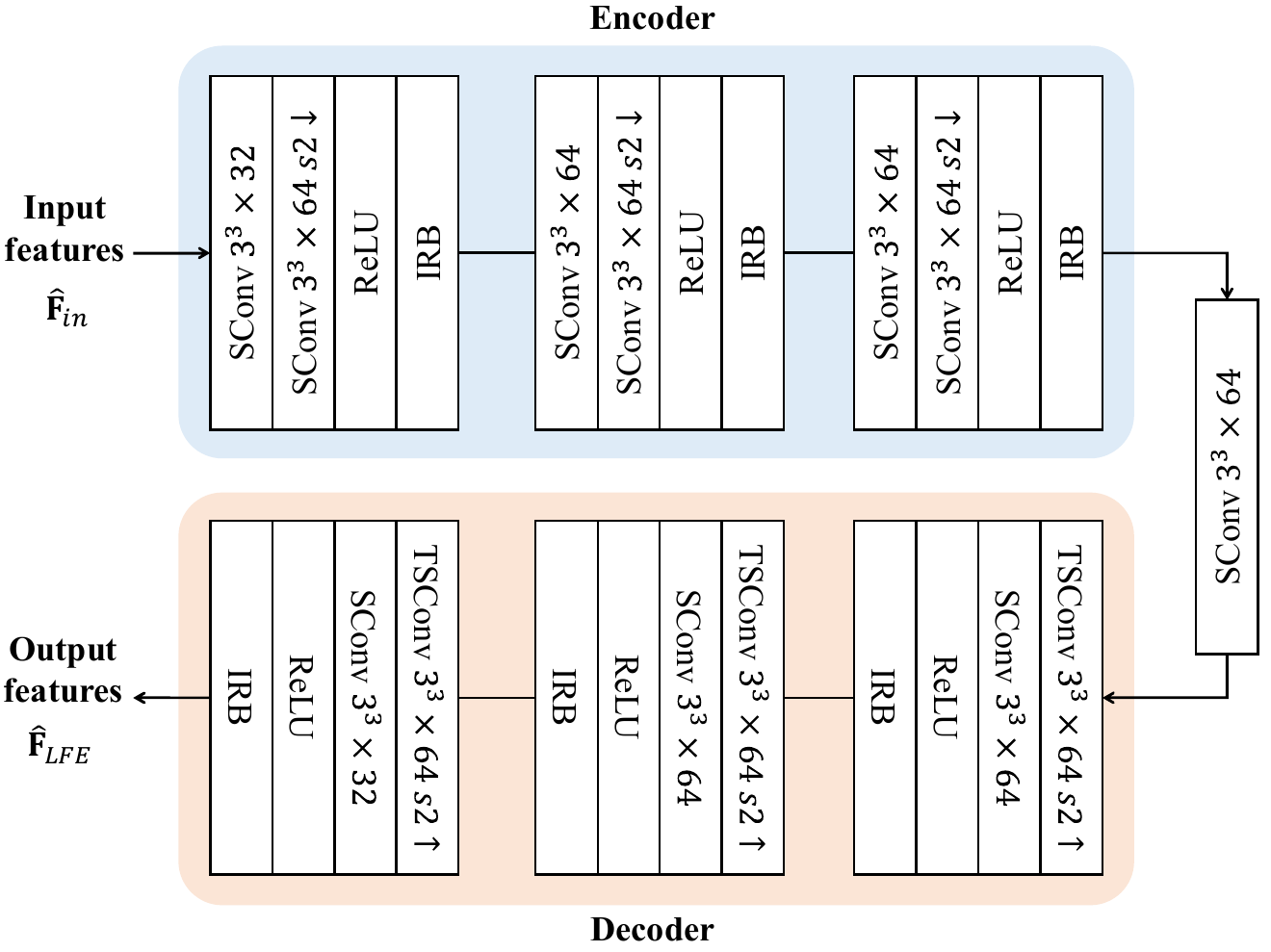}}
  \caption{{\bf Low-Frequency Embedding using an Autoencoder}. The features are progressively downsampled by three Resolution Downsamplers at the encoder for information aggregation and accordingly upsampled by three Resolution Upsamplers at the decoder to recover the resolution. Each Resolution Downsamplers/Upsampler has four layers. The encoder and decoder are symmetrical: the downdsampling layer uses SConv $k^3\times C$ $sX\downarrow$ and the upsampling layer uses TSConv $k^3\times C$ $sX\uparrow$.  $X$ is the sampling stride. $k$ and $C$ are the same as other layers without re-sampling. We set $C = 64$ at the bottleneck, assuming that low-frequency components are sufficiently embedded using such a setting.}
  \label{fig:VAE_structure}
\end{figure}

{\it Low-Frequency Embedding (LFE).} We apply a classical autoencoder shown in Fig.~\ref{fig:VAE_structure} to aggregate low-frequency  information spatially. At the encoder, input features $\hat{\bf F}_{in}$ first go through three Resolution Downsamplers for low-frequency information aggregation. Each Resolution Downsampler consists of a convolutional layer, a convolutional downsampling layer, a ReLU-based activation layer, and an IRB layer. For the convolutional downsampling layer, the sparse convolution  with a stride of 2 is applied.  As a result, at the bottleneck, the input $\hat{\bf F}_{in}$ is squeezed into $\frac{1}{8}\times\frac{1}{8}\times\frac{1}{8}$ of its original spatial size with 64 channels as exemplified. The decoder then symmetrically applies the Resolution Upsampler to upscale features using transposed sparse convolutions ``TSConv''  for the generation of output features $\hat{\bf F}_{LFE}$.

{\it High-Frequency Embedding (HFE).} Previous Low-Frequency Embedding leverages sparse convolution-based re-sampling to capture low-frequency information. 
To further characterize spatial variations from high-frequency features, another High-Frequency Embedding is devised in parallel to the Low-Frequency Embedding, as illustrated in Fig.~\ref{fig:CARNet_framework}.  
For simplicity, average pooling and upsampling operations are consecutively conducted upon the input $\hat{\bf F}_{in}$ to generate $\hat{\bf F}_{up}$; and then 
$\hat{\bf F}_{up}$ is subtracted with  $\hat{\bf F}_{in}$ to derive the high-frequency information $\hat{\bf F}_{HFE}$.  
The average pooling operation applies the sparse convolution SConv with a stride of $S$ and $k\times k\times k$ kernel, i.e., $ \hat{\bf F}_{avg}$ = $SpaseAvgPooling(\hat{\bf F}_{in}, k, S)$, and the upsampling operation applies with the transposed sparse convolution TSConv with a stride of $S$ accordingly, i.e., $\hat{\bf F}_{up} = SparseUpsampling(\hat{\bf F}_{avg}, k, S)$. It then leads to 
\begin{equation}
    \hat{\bf F}_{HFE} = \hat{\bf F}_{in} - \hat{\bf F}_{up}.
\end{equation} In this work, we set $k=3$ and $S=2$. 

We subsequently concatenate the outputs of Low-Frequency Embedding and High-Frequency Embedding subbranches, i.e., $\hat{\bf F}_{LFE}$ and $\hat{\bf F}_{HFE}$, and fuse them together using a Direct Spatial Embedding unit to produce the frequency-dependent feature tensor $\hat{\bf F}_{FDE}$.

{\bf Feature Fusion for MPSOs.} At the end of the two-stream network, both $\hat{\bf F}_{FDE}$ and $\hat{\bf F}_{DSE}$ generated from two branches are concatenated and processed using another residual Direct Spatial Embedding unit to produce a set of MPSOs as the distortion approximations of the compressed PCA.

\subsection{MPSOs Combination}
\label{subsect:inloop_filter_guided}
In the MPSOs Combination phase, we propose to linearly weigh these derived MPSOs to estimate the final sample offsets for distortion compensation. 
As seen in Fig.~\ref{fig:CARNet_framework}, after fusing the latent features from two streams, a set of MPSOs defined as ${\bf R}= [{\bf d}_0, {\bf d}_1, \cdots, {\bf d}_{H-1}]$,  are generated. Figure~\ref{fig:CARNet_framework} exemplifies three MPSOs, dubbed ${\bf d}_0$, ${\bf d}_1$, and ${\bf d}_2$. The final estimated distortion ${\bf D}_{est}$ is derived by combining these MPSOs linearly using  coefficients $A = {a_0, a_1, \cdots, a_{H-1}}$ as follows:
\begin{equation}
    {\bf D}_{est} = a_0{\bf d}_0 + a_1{\bf d}_1 + \dots + a_{H-1}{\bf d}_{H-1}.
    \label{eq:D_est}
\end{equation}
Then, the problem is how to effecitively approximate these linear weighting coefficients $A = {a_0, a_1, \cdots, a_{H-1}}$ for the square error minimization defined in \eqref{eq:carnet_alf}.

Let $\bf F$ and $\hat{\bf F}$ be the original and compressed point cloud attributes, respectively. The attribute compression distortion is:
\begin{equation}
    {\bf D} = {\bf F} - \hat{\bf F}. \label{eq:compression_disortion}
\end{equation} By combining \eqref{eq:D_est} and \eqref{eq:compression_disortion}, the problem in \eqref{eq:carnet_alf} is rewritten as
\begin{equation}
	\arg\min_{A} \left\| {{\bf D} - {\bf D}_{est}} \right\|_2^2.
\end{equation}
We then can approximate $A$ through the use of LSE optimization, i.e.,
\begin{equation}
    A=[a_0, a_1, \dots, a_{H-1}] = ({\bf R}^T{\bf R})^{-1}{\bf R}^T{\bf D},
\end{equation}
where $ {\bf R} = [{\bf d}_0, {\bf d}_1, \dots, {\bf d}_{H-1}]$ stacks vectorized MPSOs and {\bf D} is available at the encoder.

These linear weighting coefficients $a_i,i\in[0, H-1]$ are then explicitly signaled into the compressed bitstream. We can apply the proposed MPSOs Combination mechanism  either at  the frame level of the whole point cloud frame or at the block level by slicing the point cloud frame into fixed-size block patches. This work gives the frame-level example of $a_i$ signaling. 

To encode $a_i,i\in[0, H-1]$, we first scale and clip each $a_i$ to a predefined integer range and then encode it using fixed-length entropy codes. As will be shown subsequently, three weighting coefficients are used in our implementation because of the outstanding BD-Rate gains according to our experimental results. These three coefficients are all scaled by 128, bounded in the range of [-16, 15], and represented using 5-bits codes thereafter.

Using such a limited number of weighting coefficients has almost no impact on the bit rate consumption of compressed PCA but benefits the reconstruction quality significantly. For example, three 5-bits coefficients only consume 15 bits in total per frame; given a point cloud that has 800,000 points, the bit rate cost of these three weighting coefficients is as small as 1.9E-5 bits per point (bpp), which is negligible to the bit rate consumption of G-PCC compressed attributes.

%{\bf Preprocessing for U and V components.} 
\subsection{Chroma Filtering Using Cross-Component CARNet}

\begin{figure}[tbp] 
\centering 
    \includegraphics[width=\linewidth]{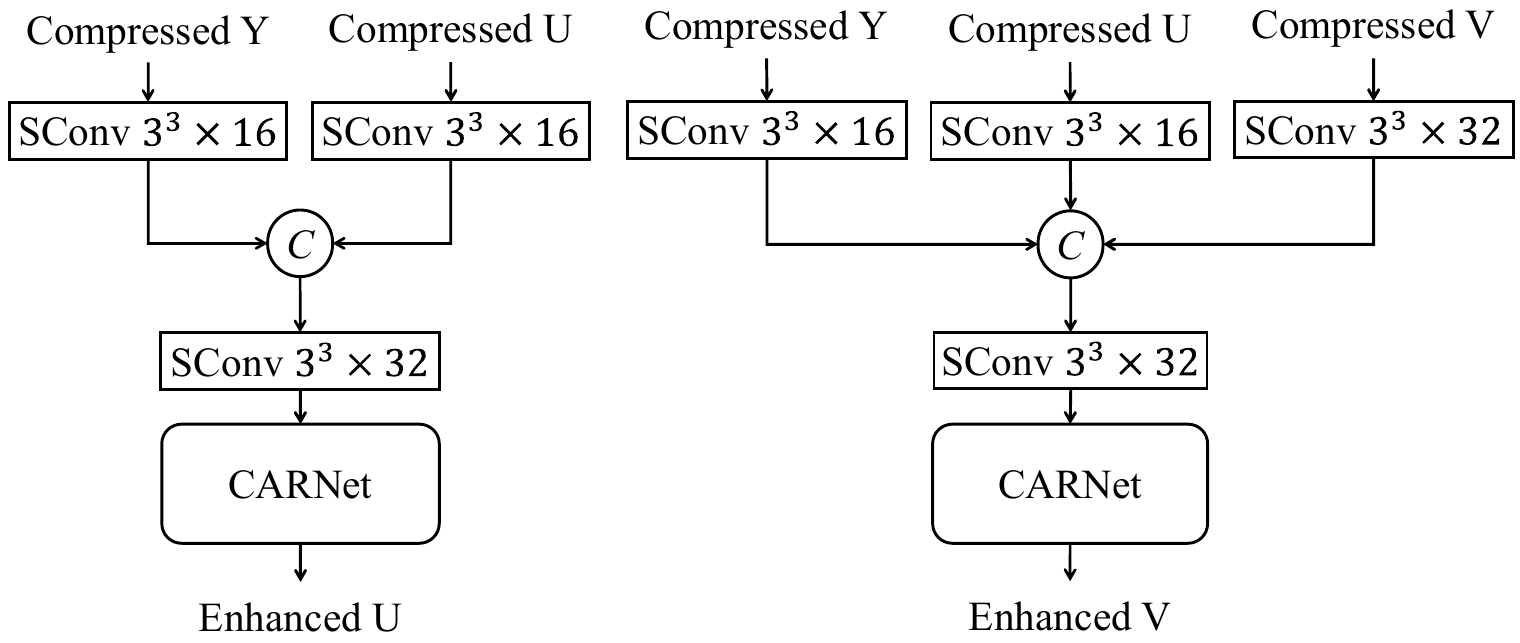}
\caption{{\bf The Cross component strategy for U and V attribute modeling}. We use the compressed Y attribute to assist the restoration of U component and use Y and U together to assist the restoration of V component.} 
\label{fig:preprocessUV}
\end{figure}

Previous sections detail the processing of Y component using the proposed CARNet.
This section extends the CARNet to process U and V components. As has been verified in previous works~\cite{lin2022nr,zhang2018enhanced}, there exist sufficient correlations across Y, U, and V components, which is so-called cross-component correlations. Such cross-component correlations are leveraged in this work for performance improvement. We propose a cascading cross-component strategy which uses the compressed Y attribute that contains rich texture details to help process the U component and both Y and U components to help process the V component. The specific network structures of U and V components are shown in Fig.~\ref{fig:preprocessUV}.  

\section{Experimental Results} 
\label{sect:experimental_results}

\subsection{Experimental Settings}
\label{subsect:experimental_settings}

%% vs G-PCC
\begin{figure*}[t] 
\centering 
\newcommand{\myfigsize}{0.215}
\newcommand{\myspace}{-6mm}
    %Y
    \includegraphics[width=\myfigsize\linewidth]{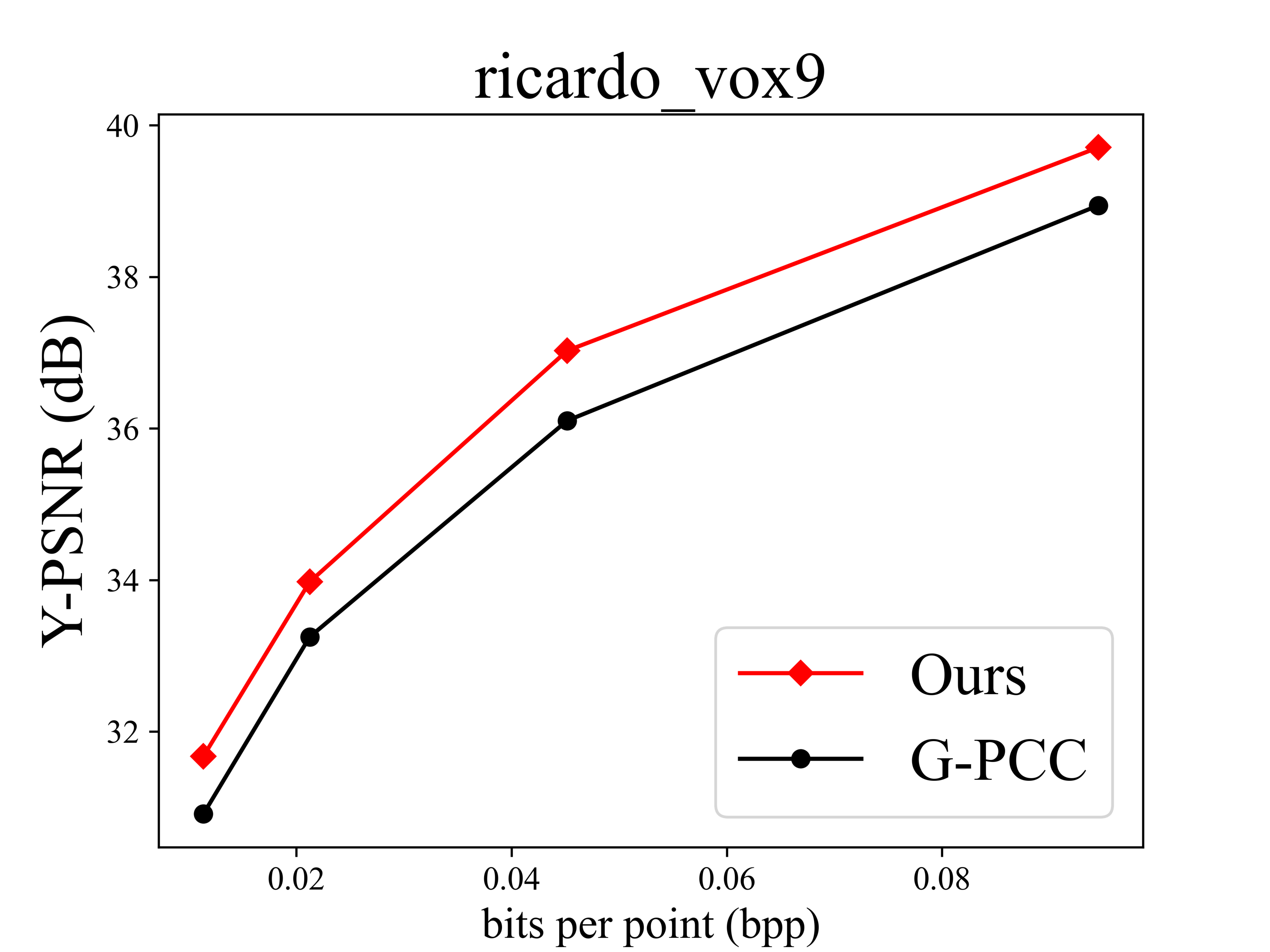} \hspace{\myspace}
    \includegraphics[width=\myfigsize\linewidth]{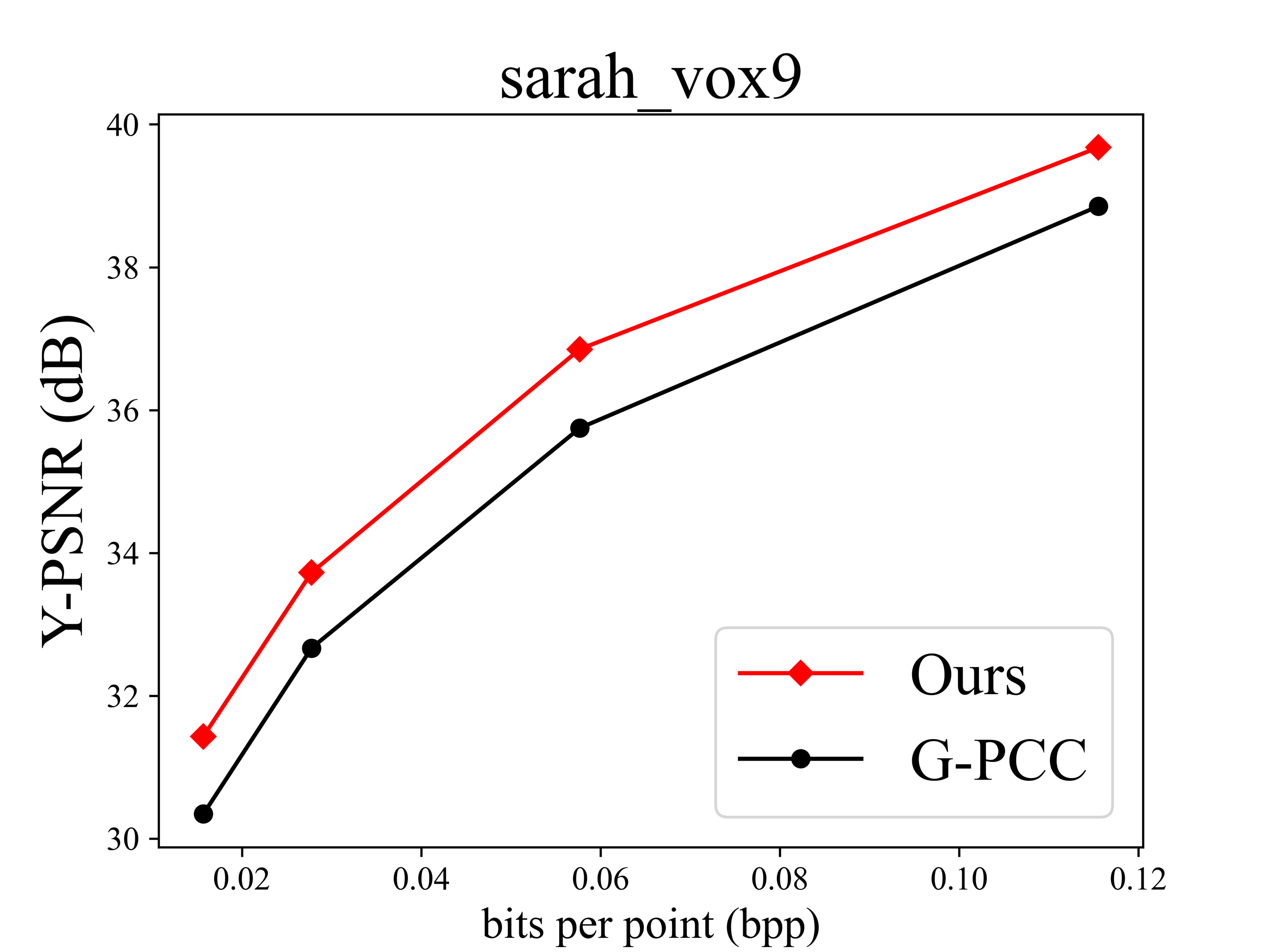} \hspace{\myspace}
    \includegraphics[width=\myfigsize\linewidth]{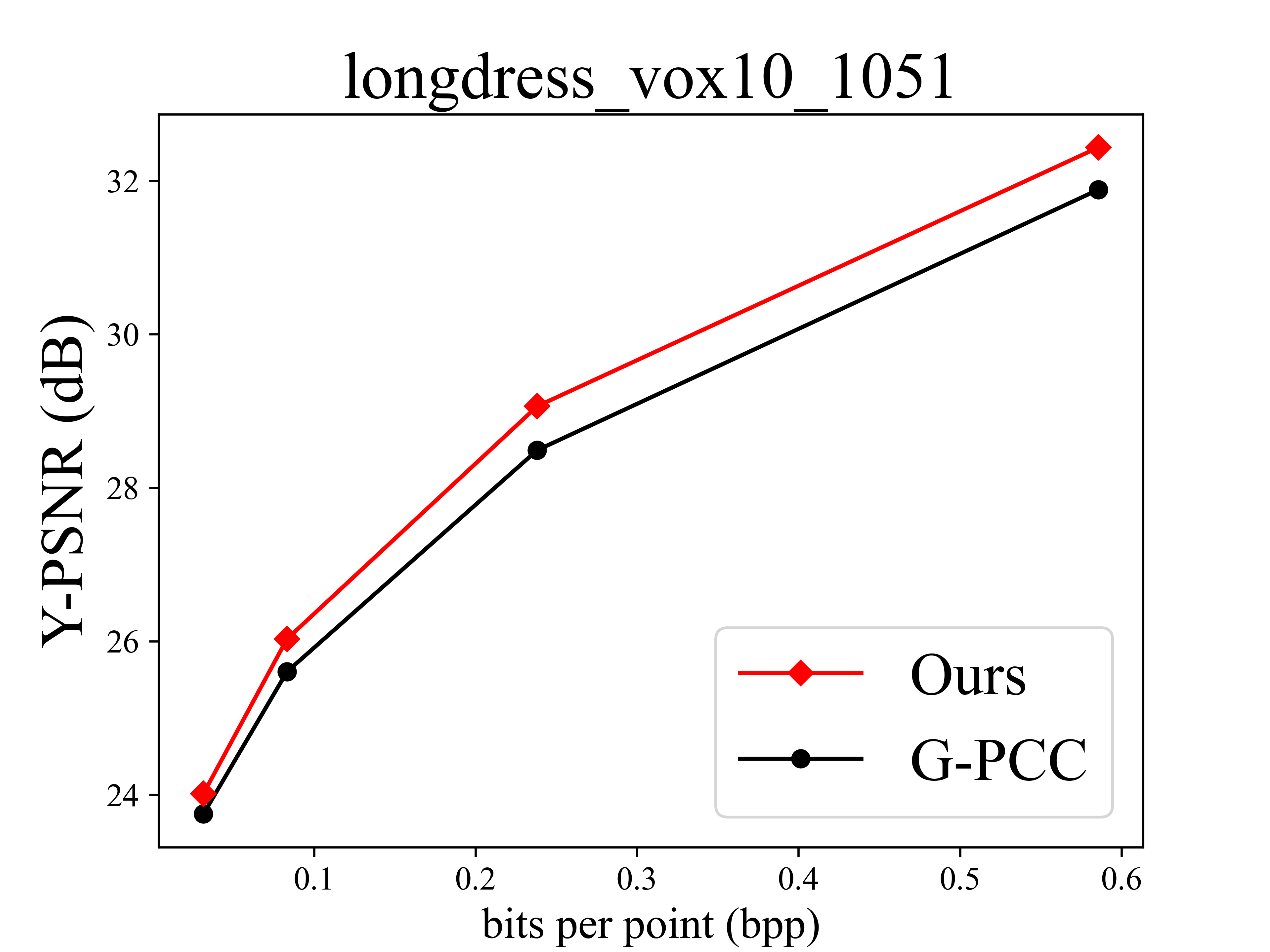} \hspace{\myspace}
    \includegraphics[width=\myfigsize\linewidth]{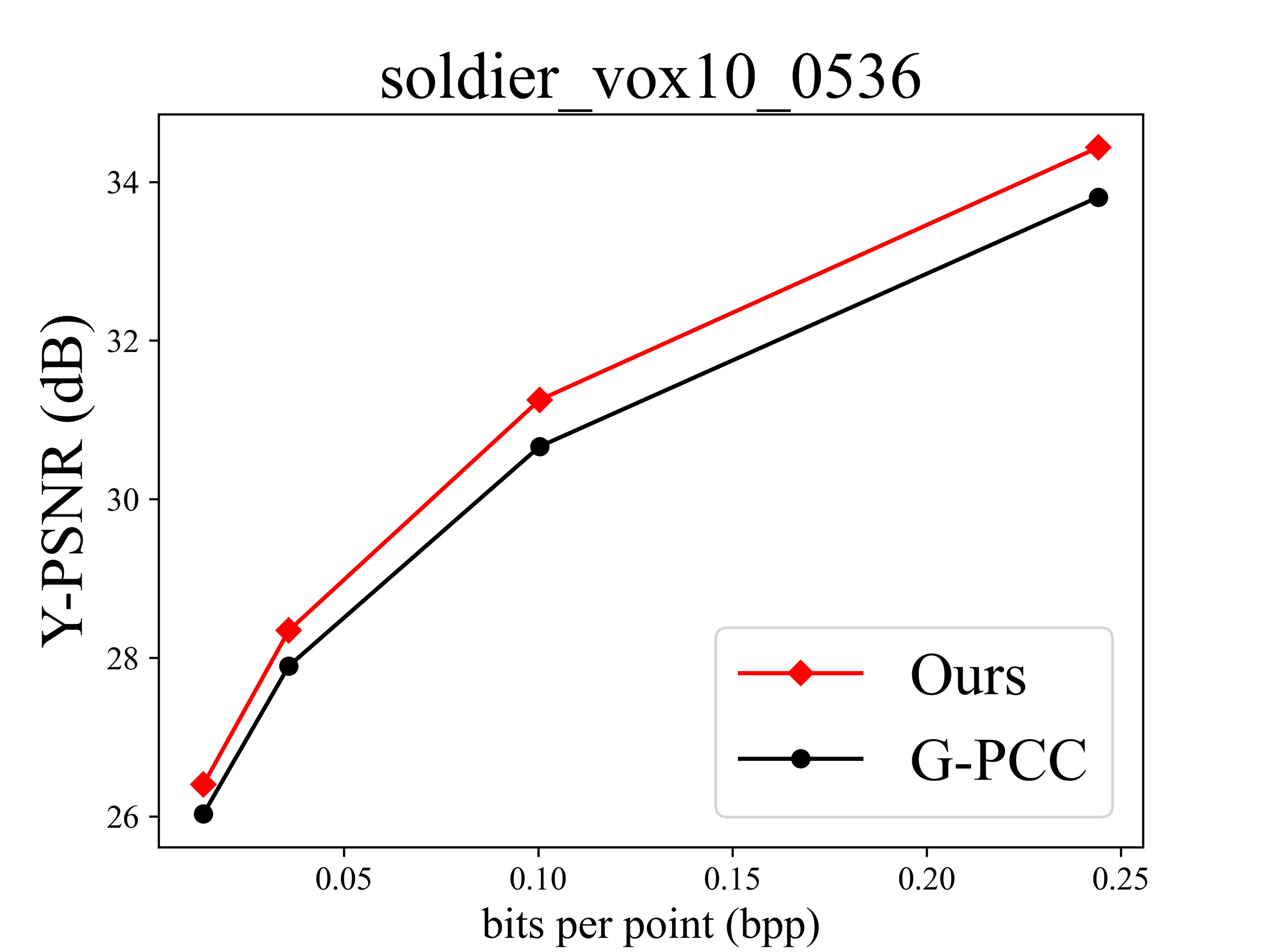}  \hspace{\myspace}
    \includegraphics[width=\myfigsize\linewidth]{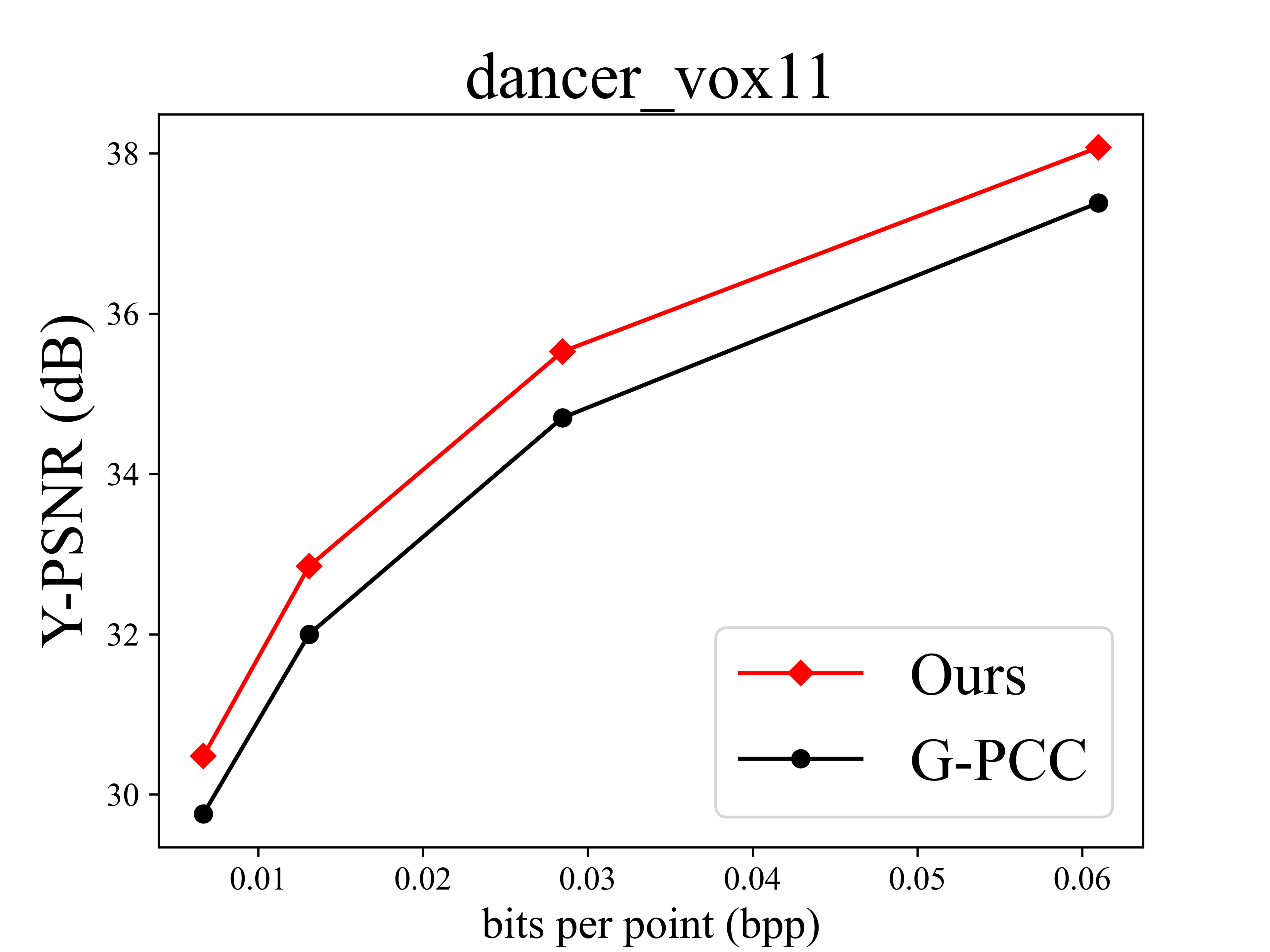}
    \\
    %YUV
    \includegraphics[width=\myfigsize\linewidth]{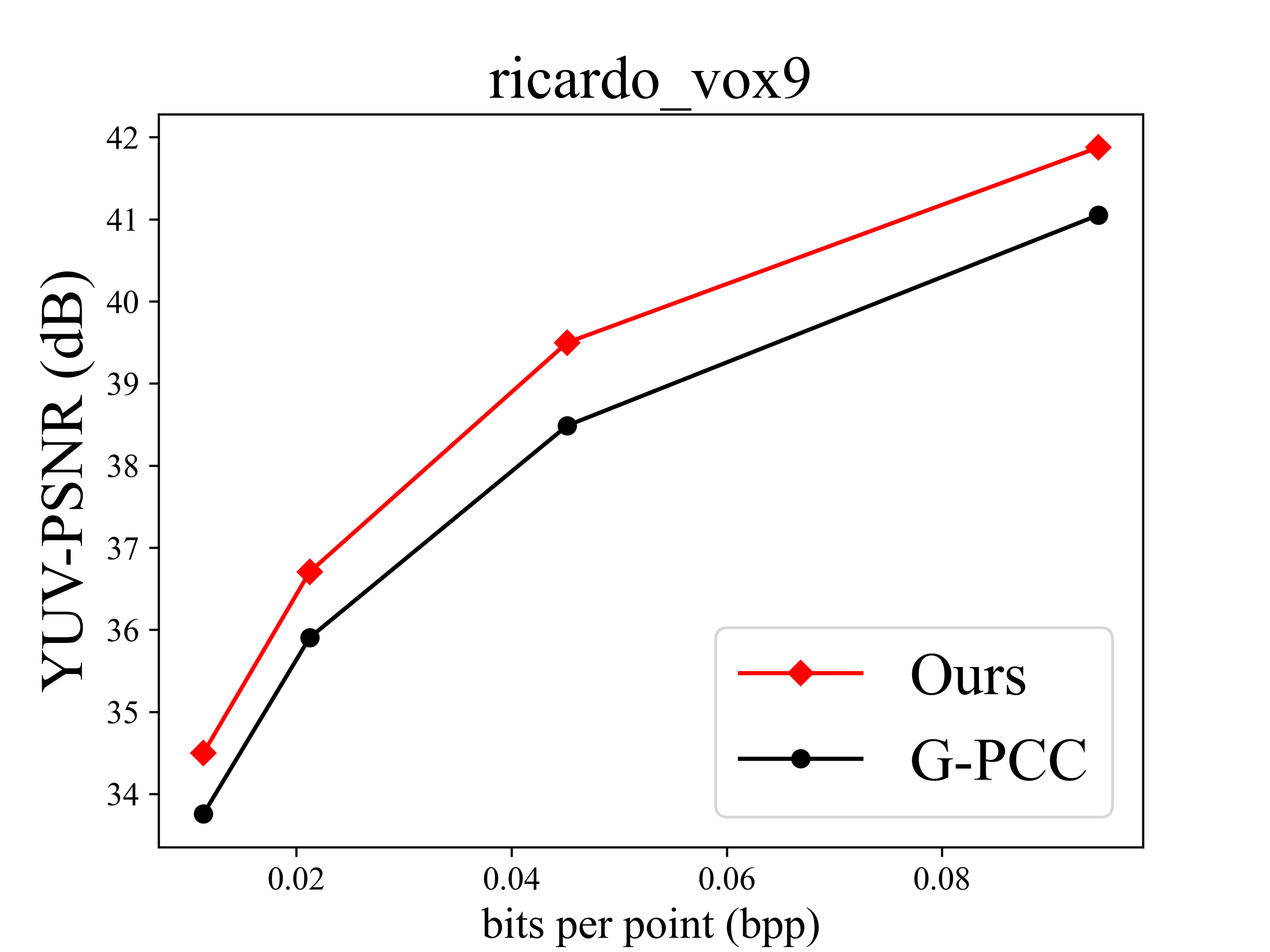} \hspace{\myspace}
    \includegraphics[width=\myfigsize\linewidth]{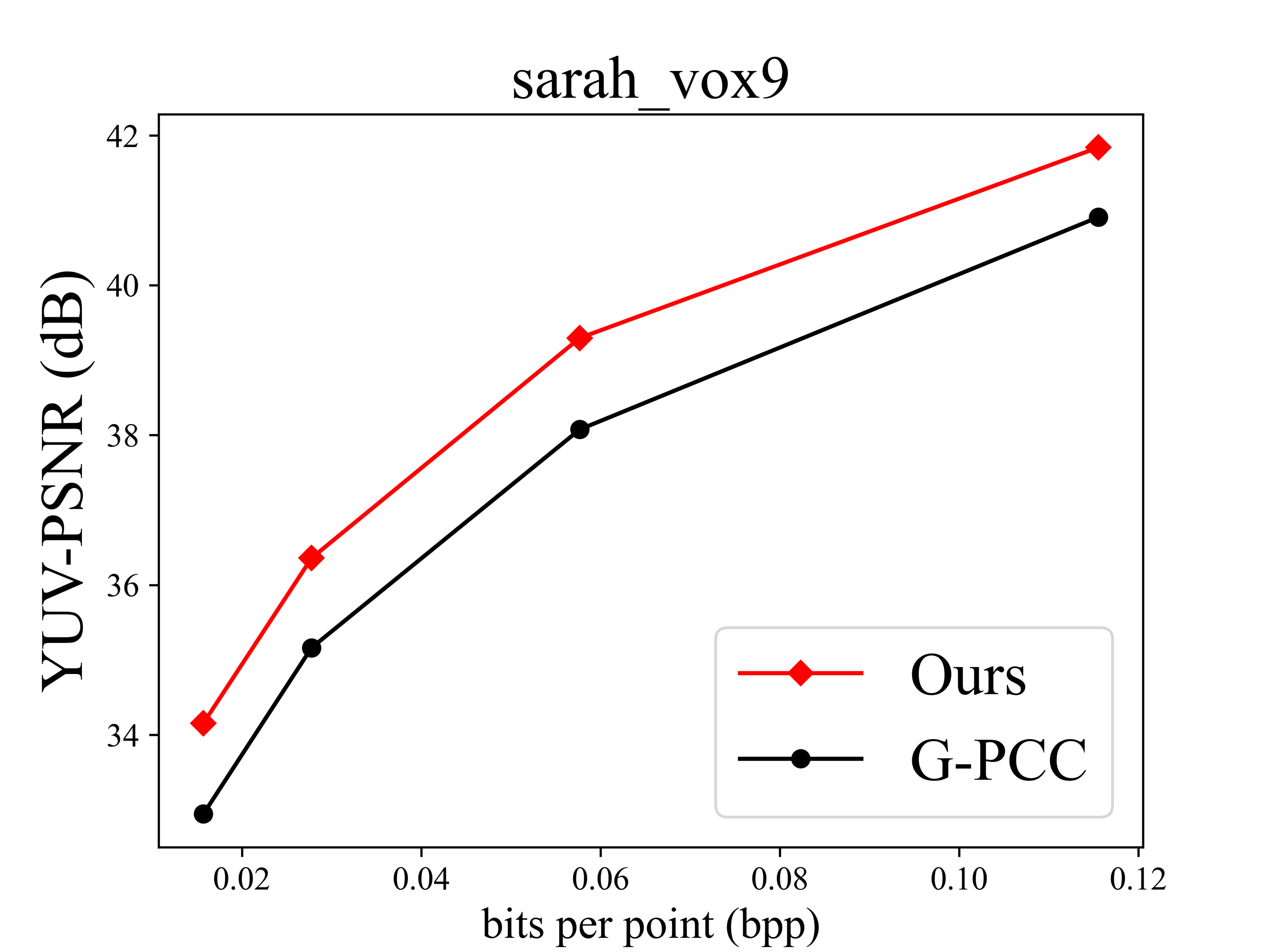} \hspace{\myspace}
    \includegraphics[width=\myfigsize\linewidth]{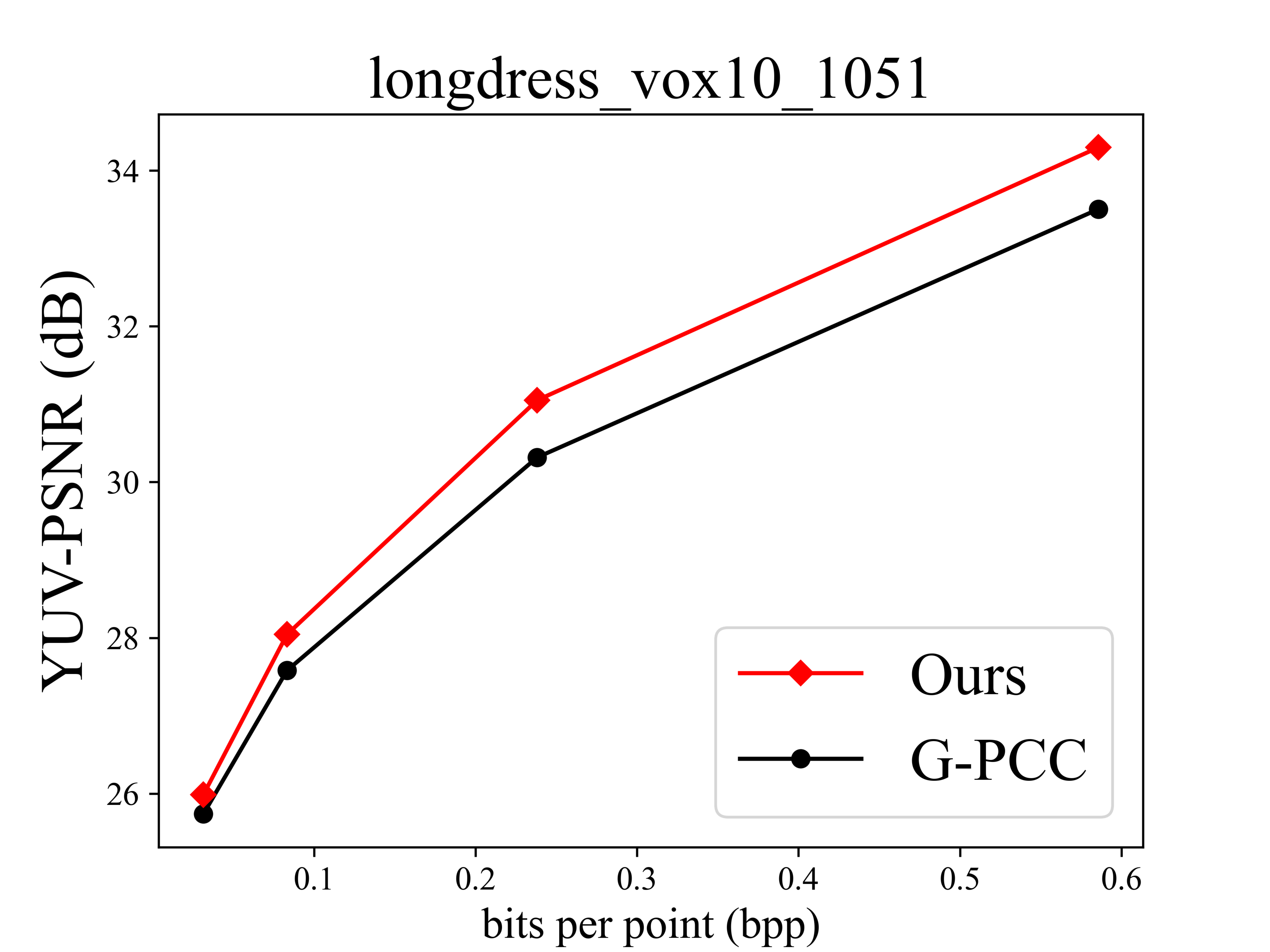} \hspace{\myspace}
    \includegraphics[width=\myfigsize\linewidth]{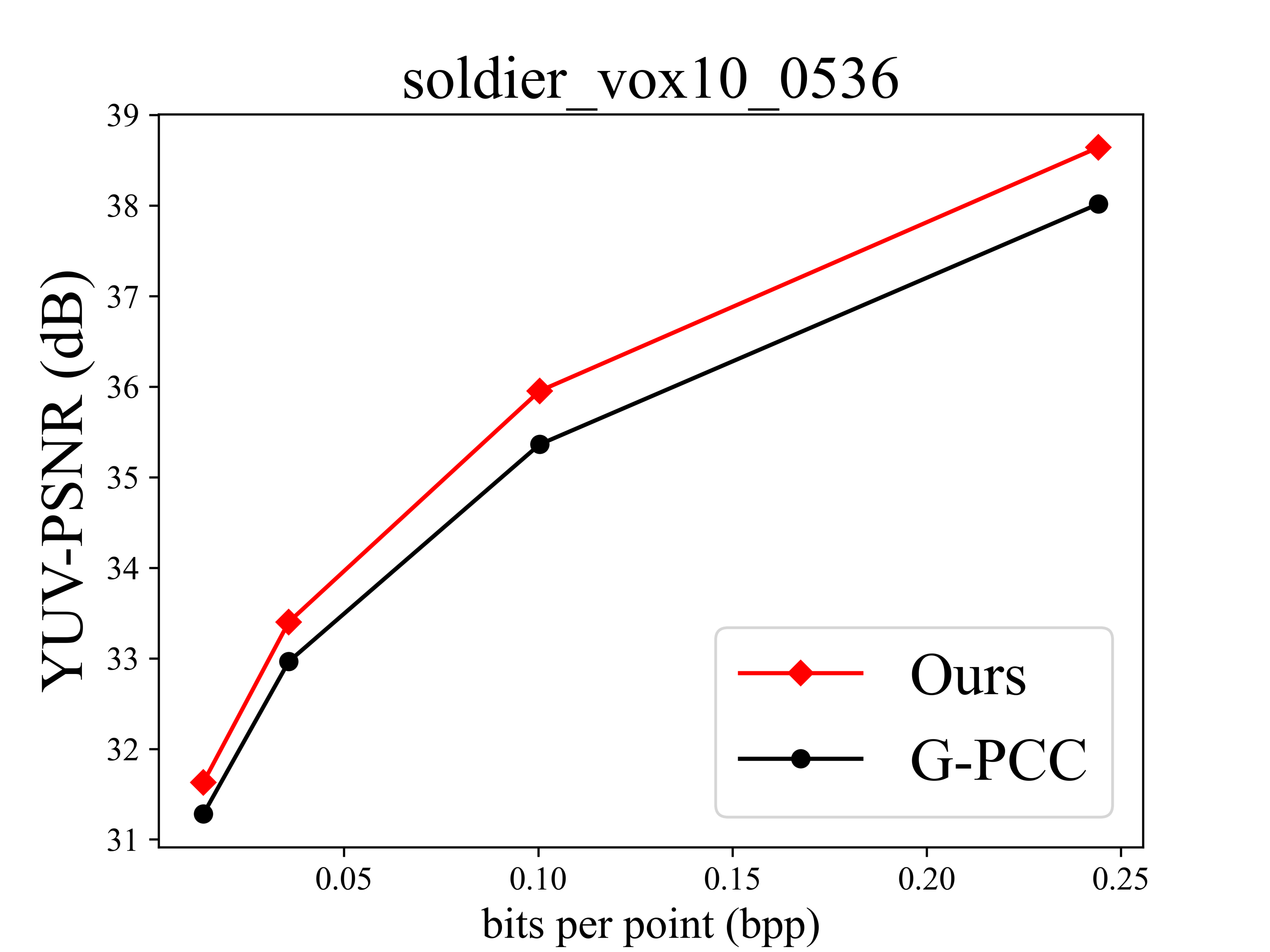}  \hspace{\myspace}
    \includegraphics[width=\myfigsize\linewidth]{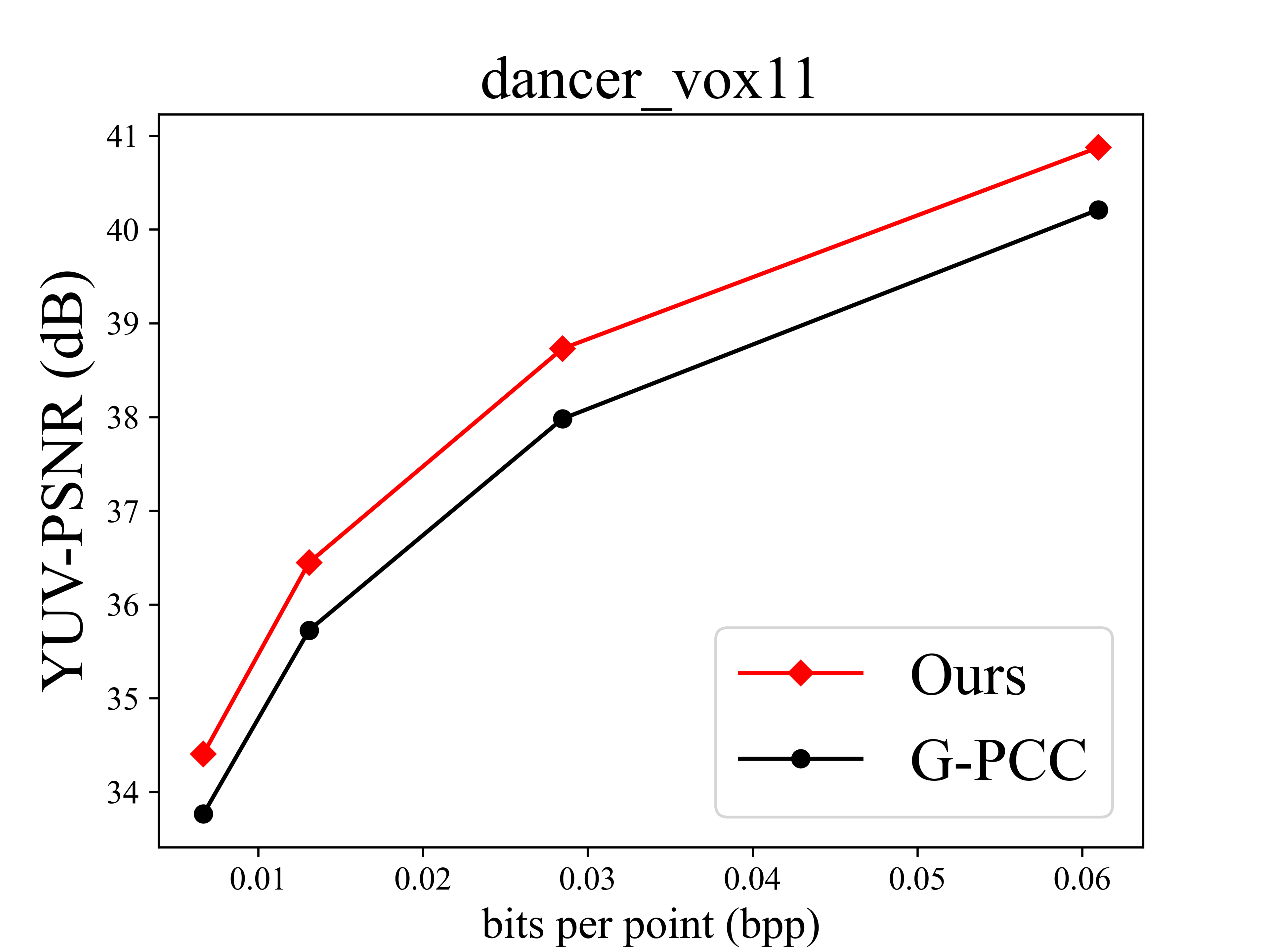}
\caption{{\bf R-D curves of the G-PCC and the proposed CARNet.} The colorized ShapeNet dataset is used to train CARNet models. R-D curves measured in Y or YUV space are both provided.} 
\label{fig:compare_RD_GPCC}
\end{figure*}

{\bf Training and Testing Dataset.}
We use the ShapeNet~\cite{chang2015shapenet} that consists of  10,000 3D models to build our training dataset. Following the method suggested by the SparsePCAC~\cite{PCAC}, we quantize the geometry coordinates of ShapeNet to 8-bit integers and randomly use color images chosen from COCO dataset~\cite{lin2014microsoft} to paint point clouds in ShapeNet to generate their color attributes. 

We compress these colorized point clouds using the latest G-PCC reference software TMC13v14\footnote{\url{https://github.com/MPEGGroup/mpeg-pcc-tmc13}} where their color attributes are processed using the default RAHT at four Quantization Parameters (QPs), including 34, 40, 46, and 51, to generate training samples. Currently, a separate model is trained for each QP value. 

It is also worth pointing out that for fair comparisons with the MS-GAT~\cite{sheng2022attribute}, our CARNet is refined using the same training samples provided by the MS-GAT. For both the MS-GAT and the proposed CARNet, we assume the G-PCC as the compression backbone and keep the same G-PCC codec configuration for comparison. Following the conventions used in the G-PCC and MS-GAT, we process the point cloud attribute in YUV domain. 

In the tests, we evaluate the proposed CARNet using 14 different point clouds widely used in standardization committees, including 5 from the Microsoft Voxelized Upper Bodies (MVUB, 9 bit), 5 from the 8i Voxelized Full Bodies (8iVFB, 10 bit), and 4 from the Owlii dynamic human mesh (Owlii, 11 bit). Notice that the training samples generated from colorized ShapeNet models are very different from the testing point clouds, revealing the generalization of the CARNet.

{\bf Training and Testing Settings.} Our project is implemented using PyTorch and MinkowskiEninge~\cite{choy20194d} on a computer with an Intel$^\circledR$ i7-8700K CPU, 32 GB memory, and Nvidia TITAN RTX GPU.  The network model is optimized by Adam and parameters $\beta1$ and $\beta2$ are set to 0.9 and 0.999, respectively. The learning rate decays from 1e-4 to 1e-5 every two epochs. We randomly initialize the model, and each model is trained on the dataset for up to 20 epochs. It takes around 70 hours for the model to converge on our platform. 

The test strictly follows the MPEG common test condition (CTC). We use bpp~({\it i.e.}, bits per point) to measure how many bits are required for each point and use PSNR~({\it i.e.}, Peak Signal-to-Noise Ratio) to evaluate the restored quality of Y, U, V components individually and YUV jointly. The quality computation of individual color components and combined YUV follows the same procedure provided in G-PCC reference software for fair comparisons. And the bits consumption for three linear weighting coefficients is also included to derive the bpp. Bj{\o}ntegaard Delta bit rate~(BD-Rate)~\cite{BDrate} is used to measure the average Rate-Distortion (R-D) performance across four QP values. 
 
\subsection{Performance Evaluation}
\label{subsect:performance_evaluation}

\begin{table}[ht]
\renewcommand{\arraystretch}{\mytablespace} 
  \centering
  \caption{BD-Rate gains of the proposed CARNet against the G-PCC (TMC13v14) measured in respective Y, U, V, and YUV spaces}
  \label{tab:compare_GPCC}%
  %\begin{scriptsize}
   \resizebox{\linewidth}{!}{
    \begin{tabular}{c|c|cccc} \hline
    {} & {Point Cloud} &  Y & U & V & YUV \\ \hline
    \multirow{5}{*}{\begin{tabular}[c]{@{}c@{}}MVUB\\ 9-bit\end{tabular}}
    & andrew & -15.30\%  & -41.90\%  & -29.08\%  & -18.36\%  \\
    & david & -18.35\%  & -34.25\%  & -34.41\%  & -23.38\%  \\
    & phil & -21.52\%  & -30.44\%  & -28.45\%  & -23.16\%  \\
    & ricardo & -19.52\%  & -34.60\%  & -23.69\%  & -22.32\%  \\
    & sarah & -22.10\%  & -33.22\%  & -33.50\%  & -25.54\%  \\ \hline
    \multirow{5}{*}{\begin{tabular}[c]{@{}c@{}}8iVFB\\ 10-bit\end{tabular}}
    & longdress & -15.94\%  & -23.02\%  & -27.68\%  & -19.20\% \\
    & loot  & -16.63\% & -23.58\% & -27.57\% & -20.15\%\\
    & redandblack & -13.74\%  & -29.92\%  & -20.05\%  & -17.38\%  \\
    & queen & -23.50\%  & -36.35\%  & -35.06\%  & -31.61\%  \\
    & soldier & -17.13\%  & -20.55\%  & -30.02\%  & -19.25\%  \\ \hline
    \multirow{4}{*}{\begin{tabular}[c]{@{}c@{}}Owlii\\ 11-bit\end{tabular}}
    & basketball\_player & -21.62\% & -18.83\% & -19.50\% & -21.19\% \\
    & exercise  & -20.41\% & -25.30\% & -26.52\% & -21.82\% \\
    & dancer & -20.87\%  & -22.56\%  & -25.44\%  & -21.74\%  \\
    & model & -17.54\%  & -32.54\%  & -38.41\%  & -22.27\%  \\ \hline
    & {\bf Average} & {\bf -18.87\%}  & {\bf -29.07\%}  & {\bf -28.53\%}  & {\bf -21.96\%}  \\ \hline
    \end{tabular}%
    }
   % \end{scriptsize}
\end{table}%

\subsubsection{Quantitative Measurement} 
Both the G-PCC and the MS-GAT are used as anchors to evaluate the proposed CARNet.

{\bf Compared with the G-PCC.} We first compare the proposed CARNet to the G-PCC anchor using its latest reference TMC13v14.  As shown in Table~\ref{tab:compare_GPCC}, the proposed CARNet outperforms state-of-the-art G-PCC by 18.87\%, 29.07\%, 28.53\%, and 21.96\% BD-Rate gains in respective Y, U, V and YUV spaces.
It is also observed that the CARNet consistently performs well for 9-bit, 10-bit, and 11-bit point clouds, although they have quite different sources and characteristics. The R-D curves are also visualized in Fig.~\ref{fig:compare_RD_GPCC}. 

\begin{table*}[t]
\renewcommand{\arraystretch}{\mytablespace} 
  \centering
  \caption{BD-Rate reduction of proposed CARNet to G-PCC and MS-GAT~\cite{sheng2022attribute}. The CARNet is finetuned using the datasets suggested in MS-GAT.}
  \label{tab:compare_MSGAT}%
   %\resizebox{\linewidth}{!}
   {
    \begin{tabular}{c|c|cccc|cccc} \hline
    \multirow{2}{*}{} & \multirow{2}{*}{Point Cloud} & \multicolumn{4}{c|}{G-PCC~(TMC13v14)} & \multicolumn{4}{c}{MS-GAT~\cite{sheng2022attribute}} \\
       %&   & BD-Rate & BD-Rate & BD-Rate & BD-Rate & BD-Rate & BD-Rate & BD-Rate & BD-Rate  \\
       &  & Y & U & V & YUV & Y & U & V & YUV \\\hline
    \multirow{5}{*}{\begin{tabular}[c]{@{}c@{}}MVUB\\ 9-bit\end{tabular}}
    & andrew & -14.10\%  & -38.67\%  & -22.72\%  & -16.76\%  & -6.53\%  & -31.79\%  & -9.19\%  & -9.24\%  \\
    & david & -18.36\%  & -38.77\%  & -28.82\%  & -23.01\%  & -9.47\%  & -31.47\%  & -18.88\%  & -14.28\%  \\
    & phil & -21.96\%  & -27.51\%  & -25.18\%  & -22.85\%  & -11.67\%  & -22.05\%  & -20.99\%  & -13.58\%  \\
    & ricardo & -18.58\%  & -31.28\%  & -22.50\%  & -20.97\%  & -10.61\%  & -23.69\%  & -12.25\%  & -12.77\%  \\
    & sarah & -22.32\%  & -36.48\%  & -27.61\%  & -25.26\%  & -11.35\%  & -27.84\%  & -16.09\%  & -14.57\%  \\ \hline 
    \multirow{3}{*}{\begin{tabular}[c]{@{}c@{}}8iVFB\\ 10-bit\end{tabular}}
    & longdress & -18.06\%  & -22.29\%  & -27.44\%  & -20.42\%  & -6.11\% & -20.39\%  & -24.17\%  & -11.53\%  \\
    & redandblack & -15.96\%  & -28.14\%  & -19.43\%  & -18.45\%  & -6.14\%  
      & -26.12\%  & -15.41\%  & -10.96\%  \\
    & soldier & -20.21\%  & -16.70\%  & -28.71\%  & -20.97\%  & -8.90\%  & -10.83\%  & -19.02\%  & -10.54\%  \\ \hline
    \multirow{2}{*}{\begin{tabular}[c]{@{}c@{}}Owlii\\ 11-bit\end{tabular}}
    & dancer & -23.14\%  & -28.67\%  & -29.76\%  & -24.64\%  & -13.05\%  & -27.03\%  & -25.42\%  & -16.12\%  \\
    & model & -19.47\%  & -38.31\%  & -40.71\%  & -24.57\%  & -9.44\%  & -35.62\%  & -34.48\%  & -15.93\%  \\ \hline
    & {\bf Average} & {\bf -19.22\%}  & {\bf -30.68\%}  & {\bf -27.29\%}  & {\bf -21.79\%}  & {\bf -9.33\%}  & {\bf -25.68\%}  & {\bf -19.59\%}  & {\bf -12.95\%}  \\ \hline
    \end{tabular}%
    }
  \label{tab:msgat}%
\end{table*}%

\begin{figure*}[t] 
\centering 
\newcommand{\myfigsize}{0.215}
\newcommand{\myspace}{-6mm}
    %Y
    \includegraphics[width=\myfigsize\linewidth]{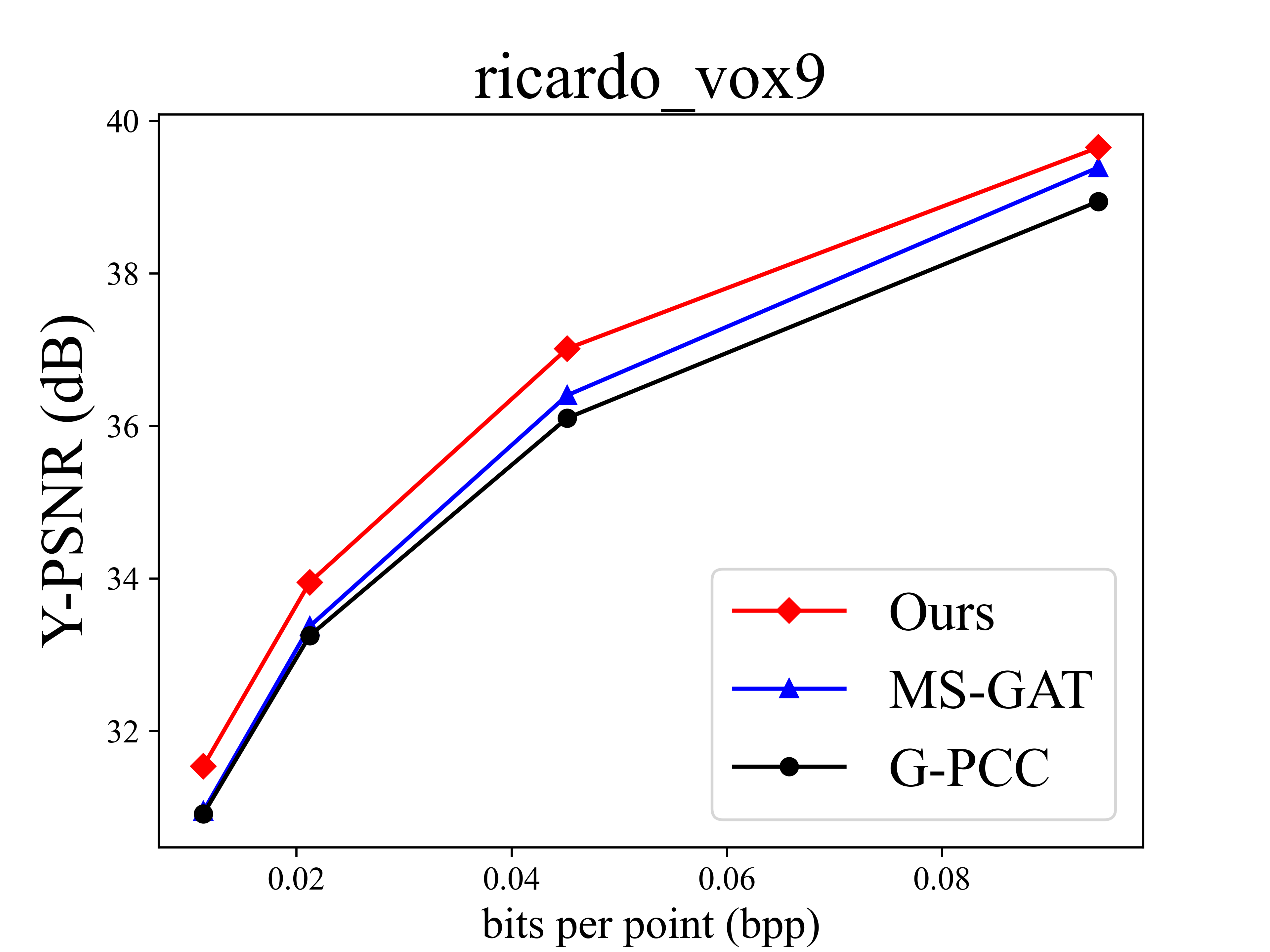} \hspace{\myspace}
    \includegraphics[width=\myfigsize\linewidth]{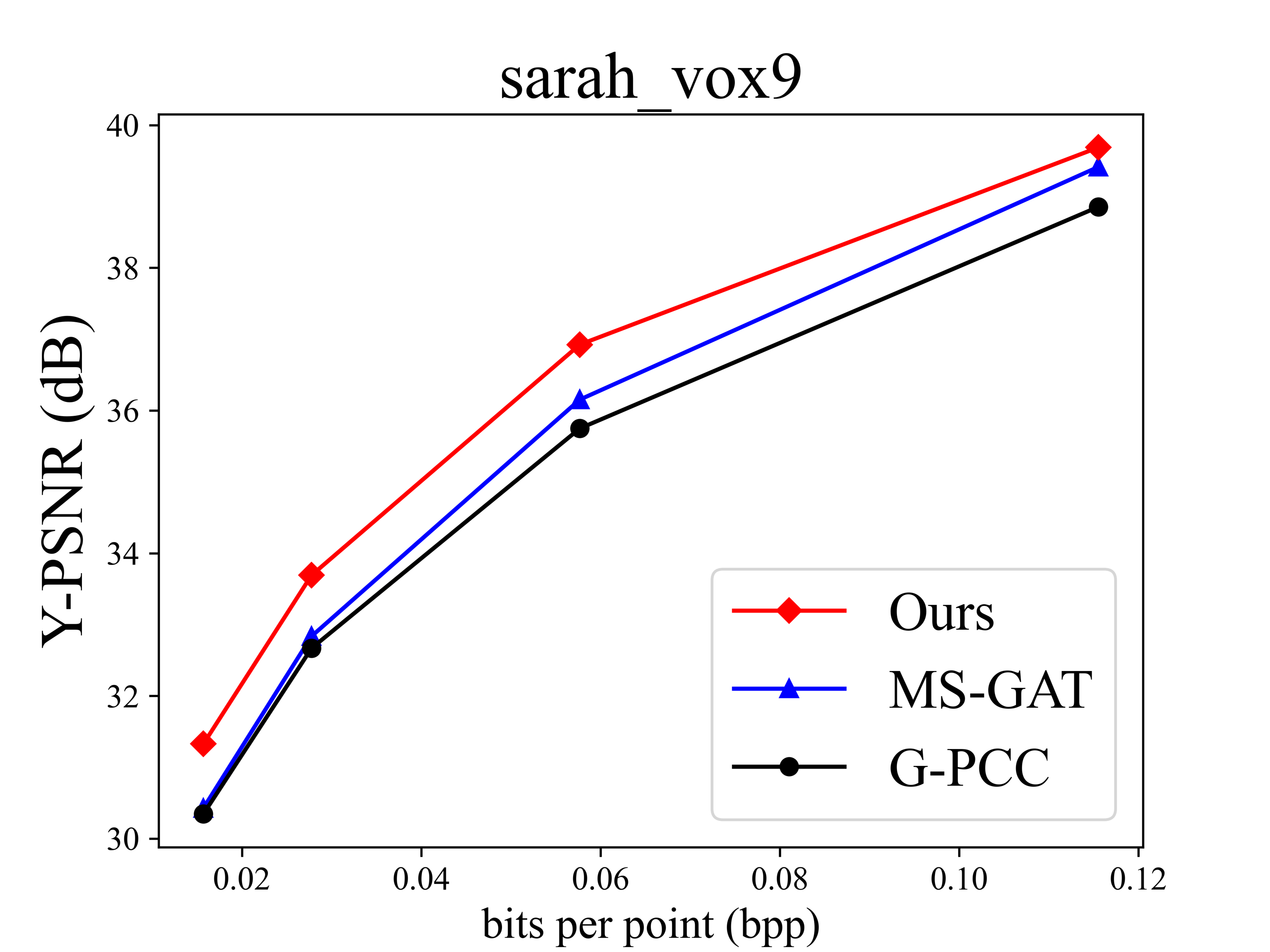} \hspace{\myspace}
    \includegraphics[width=\myfigsize\linewidth]{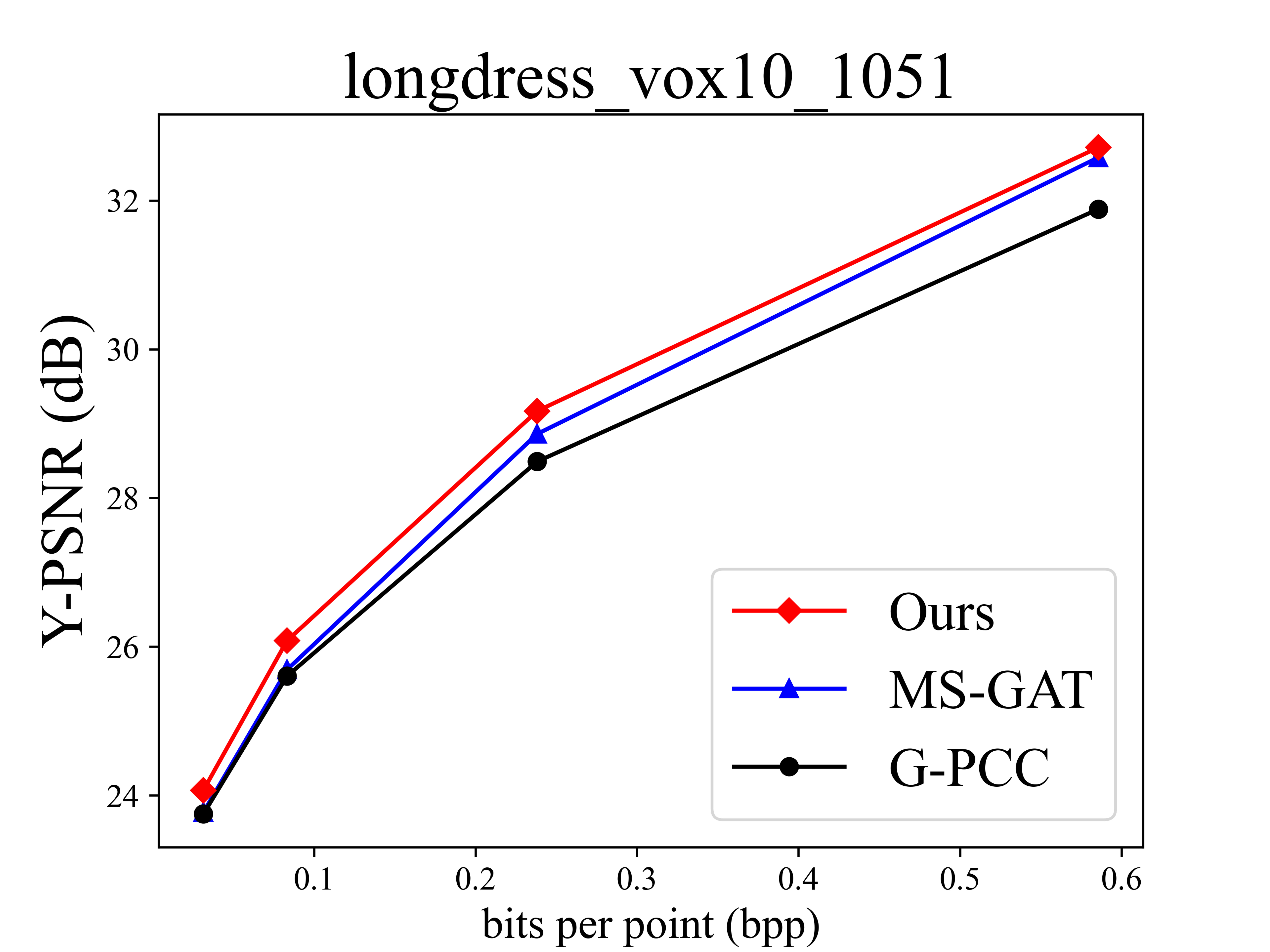} \hspace{\myspace}
    \includegraphics[width=\myfigsize\linewidth]{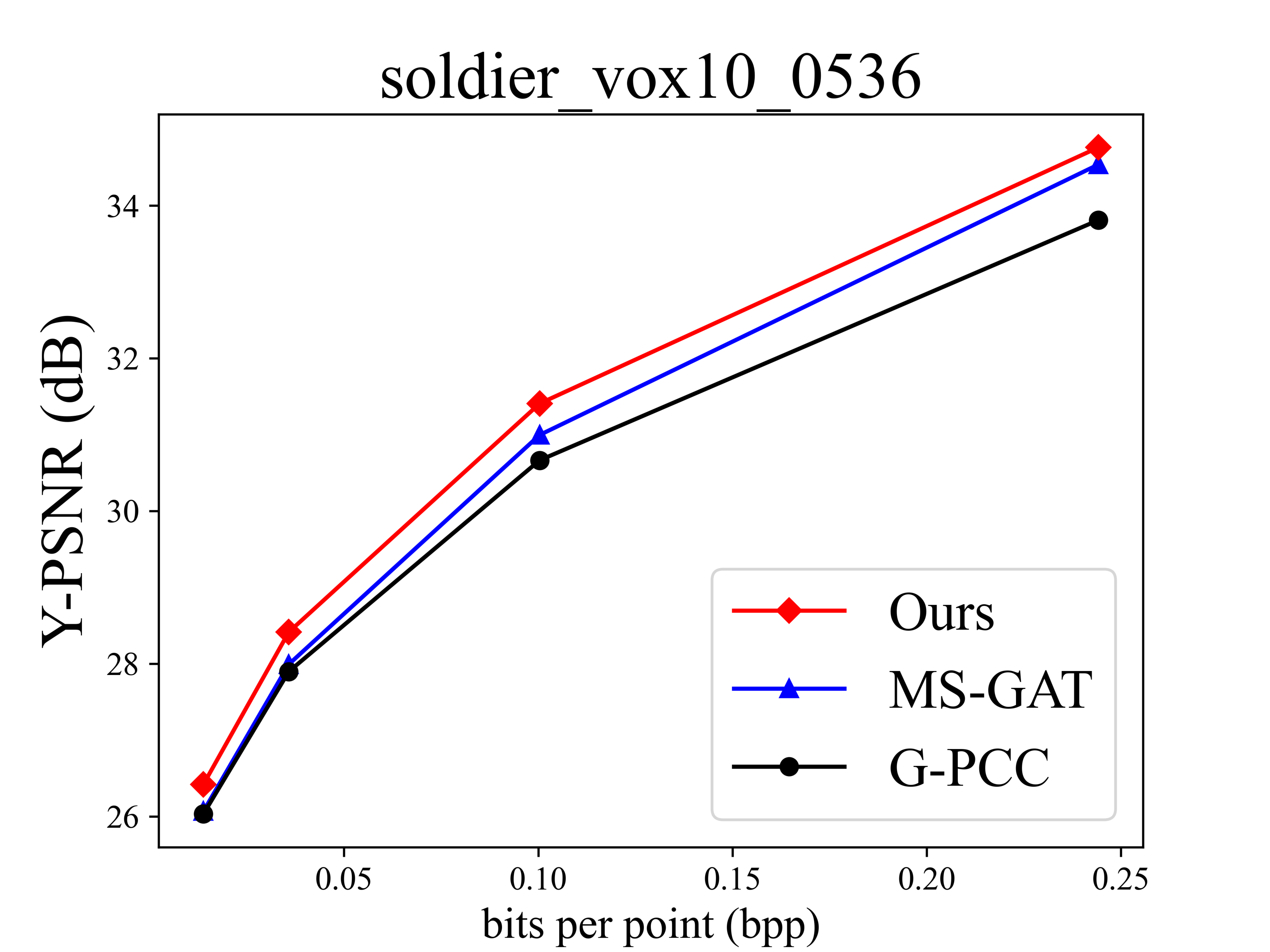}  \hspace{\myspace}
    \includegraphics[width=\myfigsize\linewidth]{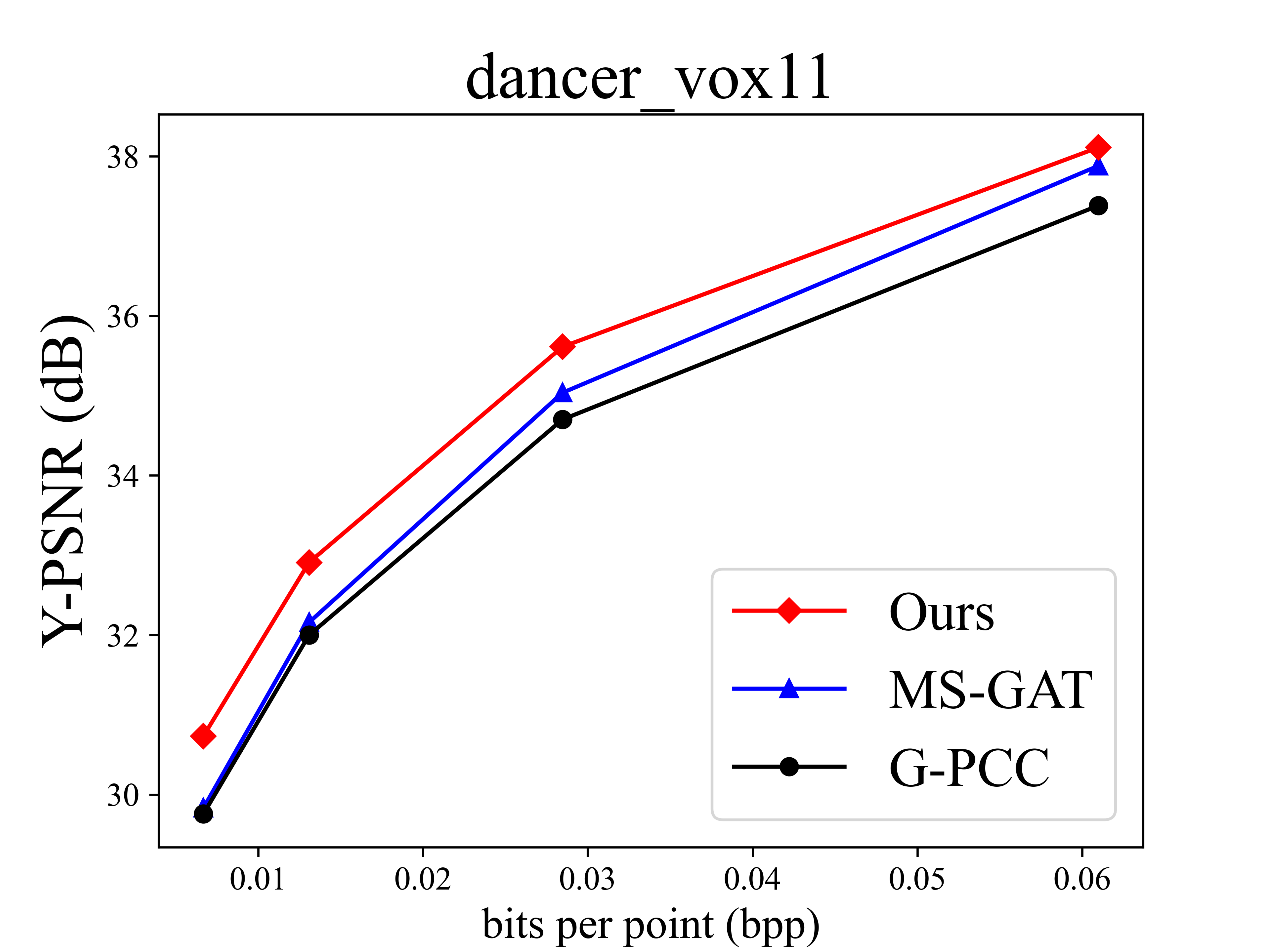} \hspace{\myspace}
    \\
    %YUV
    \includegraphics[width=\myfigsize\linewidth]{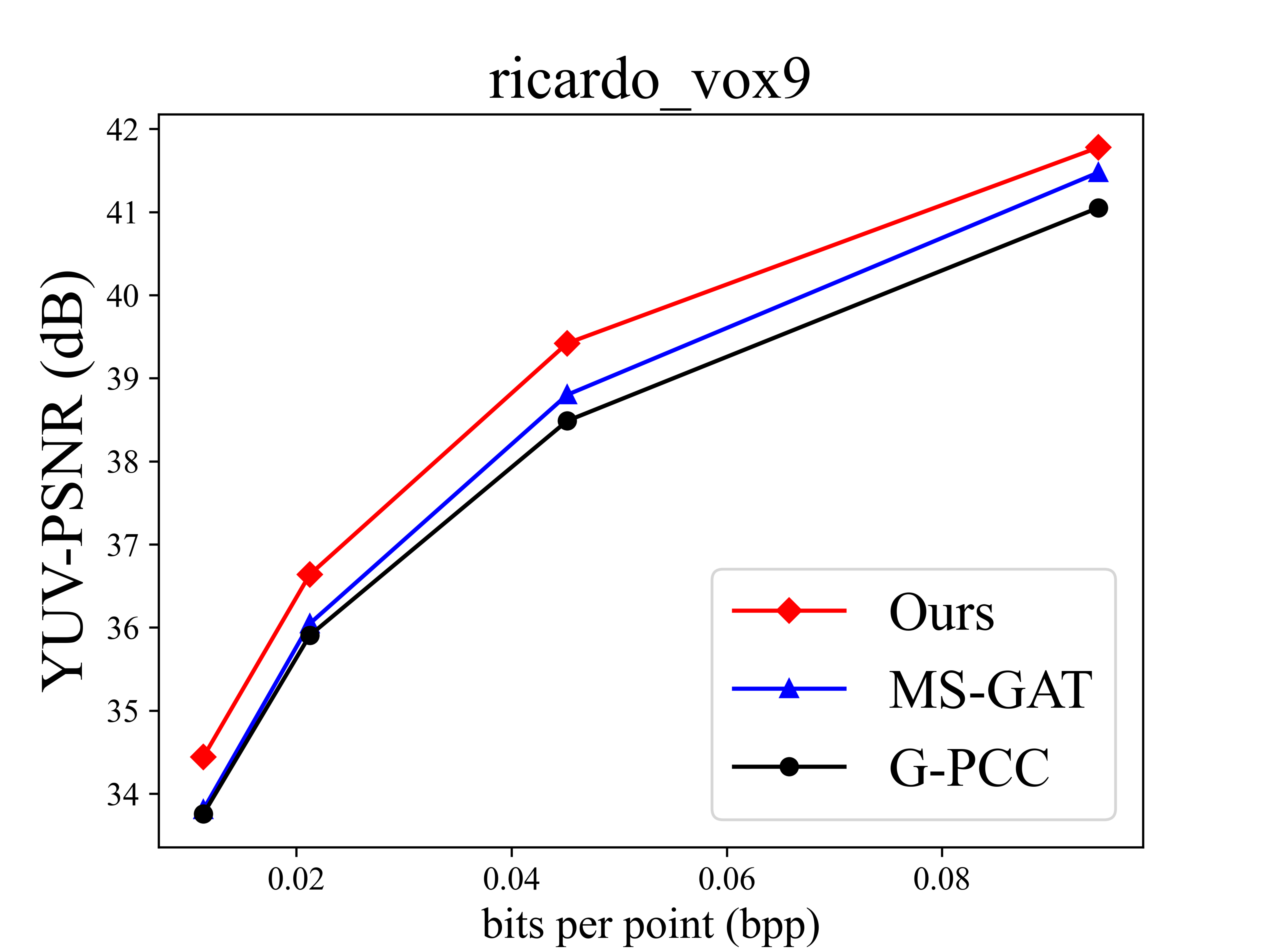} \hspace{\myspace}
    \includegraphics[width=\myfigsize\linewidth]{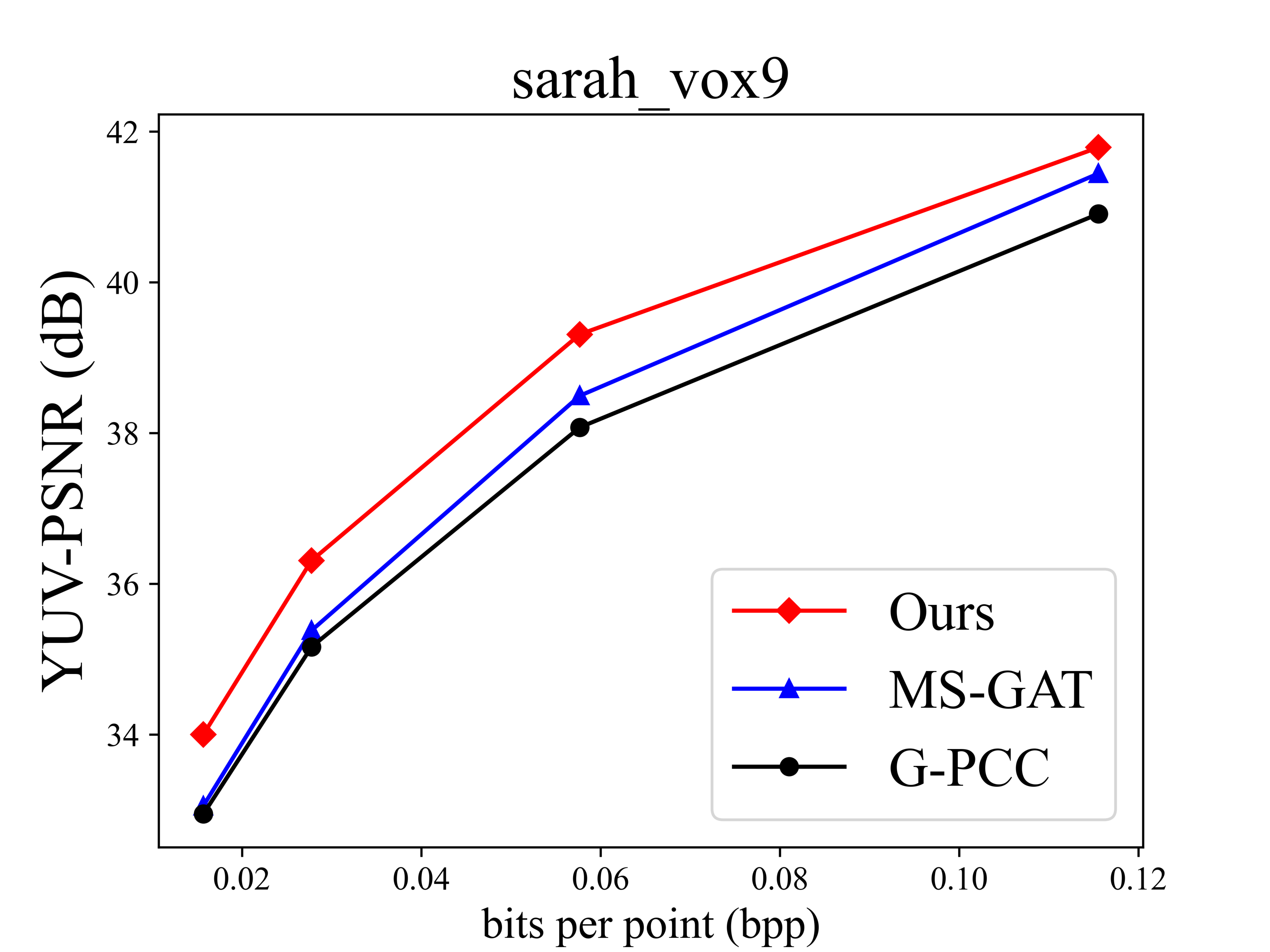} \hspace{\myspace}
    \includegraphics[width=\myfigsize\linewidth]{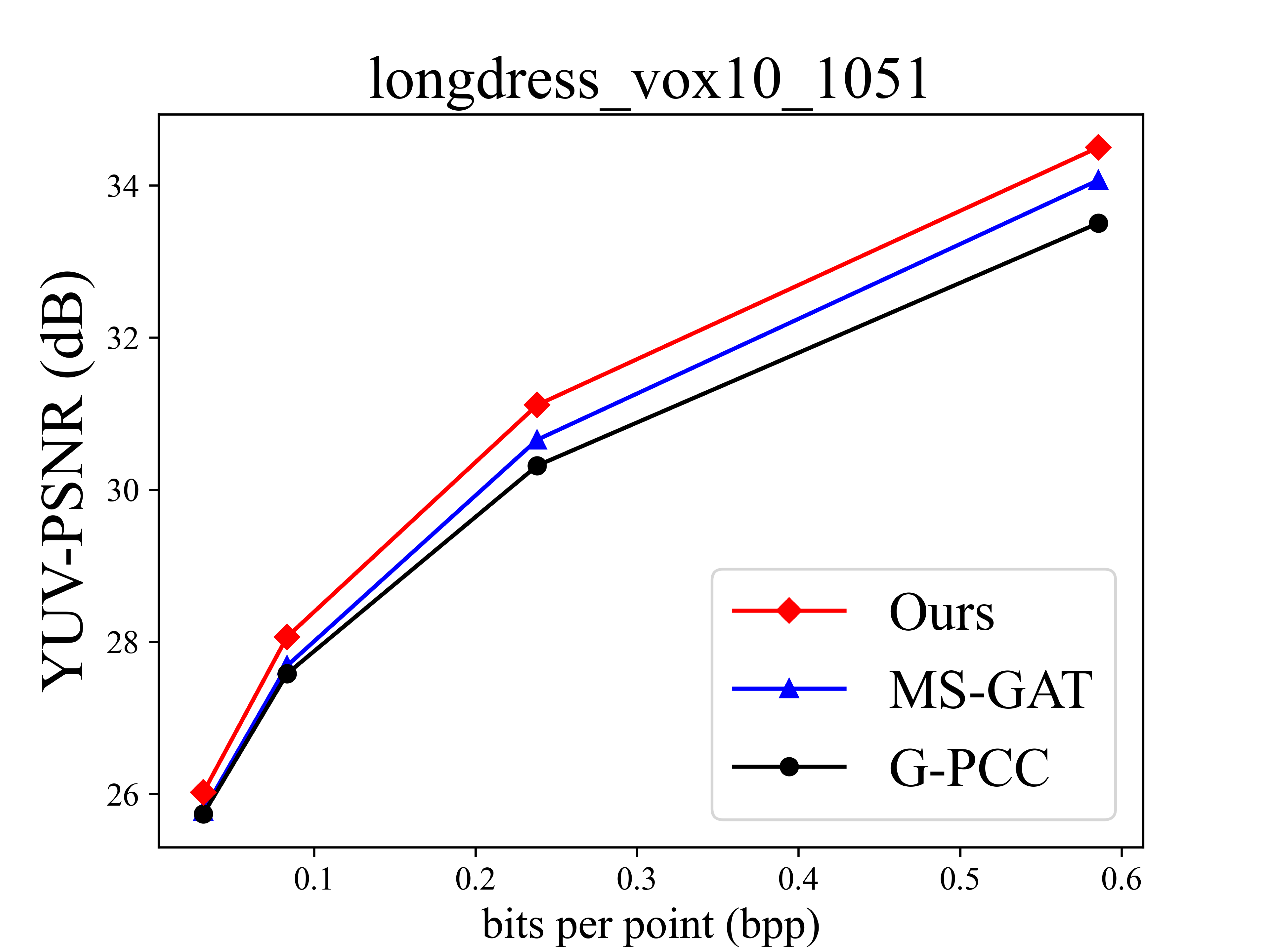} \hspace{\myspace}
    \includegraphics[width=\myfigsize\linewidth]{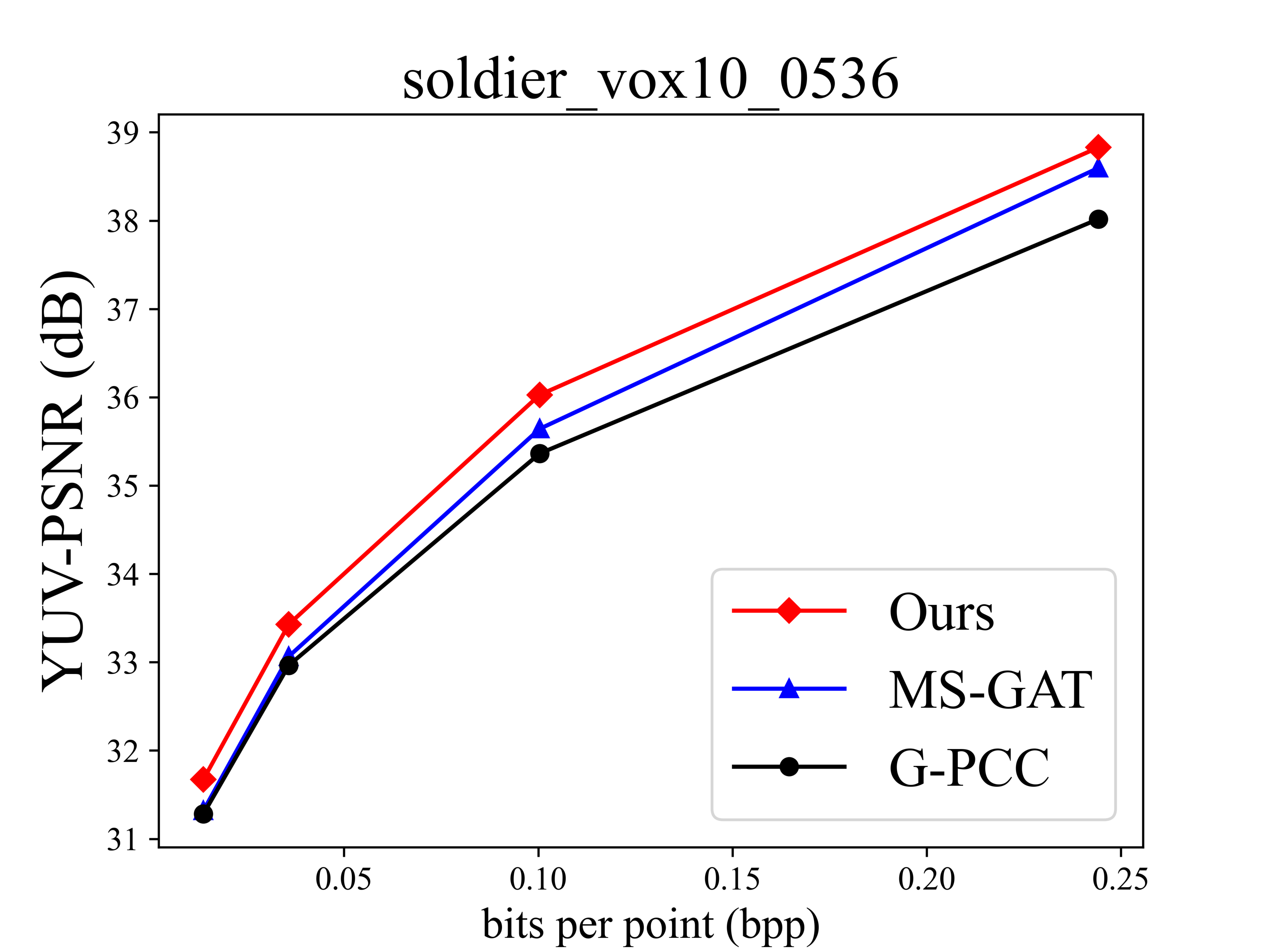}  \hspace{\myspace}
    \includegraphics[width=\myfigsize\linewidth]{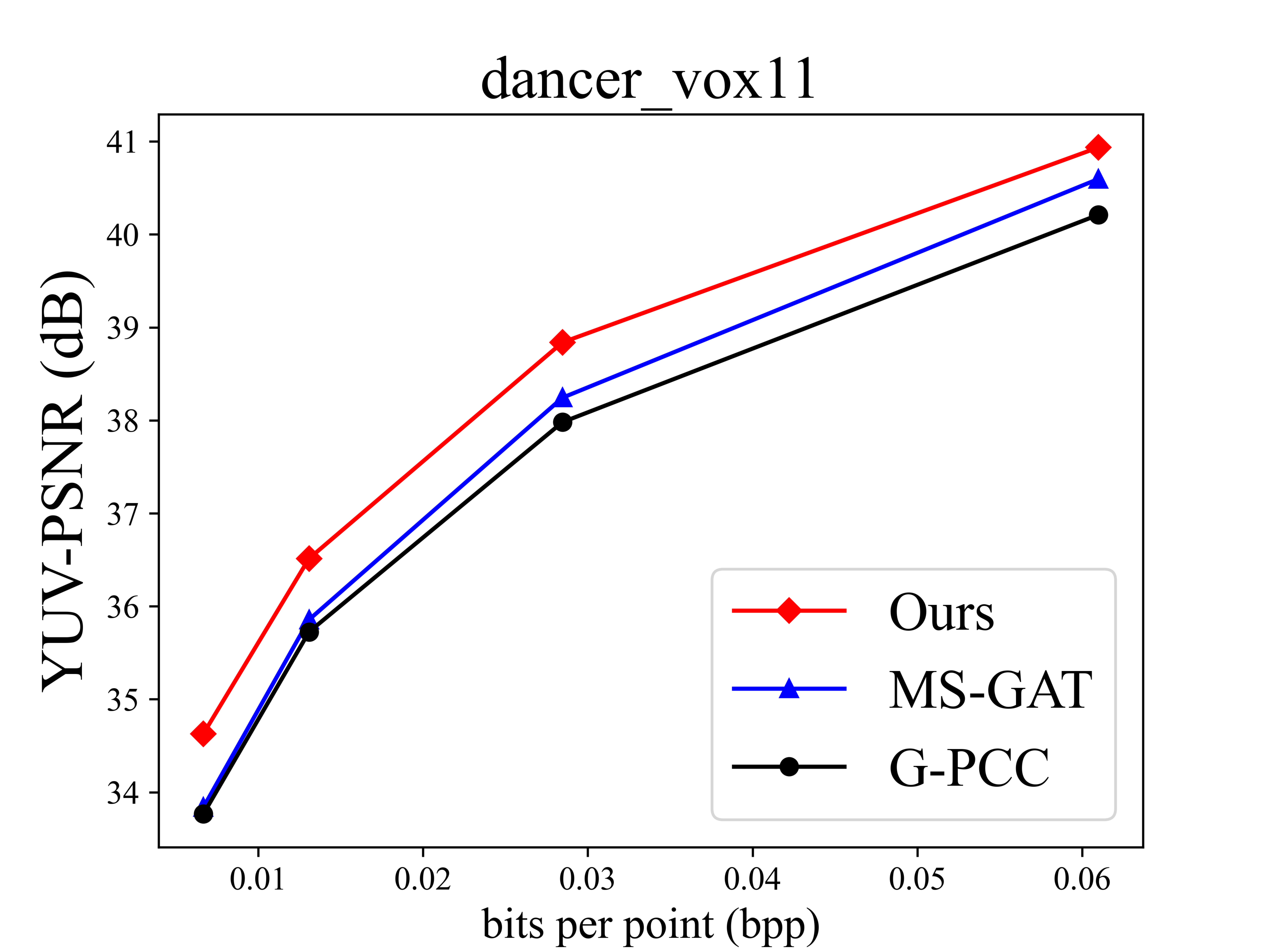}
\caption{{\bf R-D curves of G-PCC, MS-GAT, and the proposed CARNet.} The MS-GAT performance is obtained using its pre-trained models. Both Y and YUV R-D curves are provided to demonstrate the BD-Rate performance. } 
\label{fig:compare_RD_MSGAT}
\end{figure*}

%{\bf Compared with MS-GAT.}
{\bf Compared with the MS-GAT~\cite{sheng2022attribute}.}
We also compare our CARNet with the MS-GAT~\cite{sheng2022attribute} - a post-processing method for G-PCC compressed attribute artifacts removal with state-of-the-art performance. 

The MS-GAT selects five point clouds from the MPEG dataset for training (including ``basketball\_player'', ``loot'', ``exercise'', ``queen'', and ``boxer''). For fair comparisons, we build a training dataset using the same point clouds. Specifically, we partition each point cloud into 50000, 80000, and 100000 points by K-Dimensional (KD)-tree and finally obtain 648 training samples to refine the CARNet.

The performance of MS-GAT is obtained using its pretrained models. Although these MS-GAT models are trained using G-PCC reference model version 12, i.e., TMC13v12, we directly apply them on TMC13v14 compressed PCAs for testing because there is negligible performance difference between TMC13v12 and TMC13v14. We also test exactly the same point clouds used in MS-GAT.

Table~\ref{tab:compare_MSGAT} and Figure~\ref{fig:compare_RD_MSGAT} compare the enhancement gains of the MS-GAT and the CARNet. It can be seen that refined by the MS-GAT's dataset, the CARNet achieves 19.22\%, 30.68\%, and 27.29\% BD-Rate gains over the G-PCC, and 9.33\%, 25.68\%, and 19.59\% BD-Rate gains over the MS-GAT on the Y, U, and V component, respectively. For the measurement using compound YUV, the CARNet further offers superior performance to the MS-GAT, i.e., 12.95\% BD-Rate improvement.

%{\bf Visualization Comparison.}
\subsubsection{Qualitative Visualization}
\begin{figure*}[tbp] 
\centering 
    \includegraphics[width=0.98\linewidth]{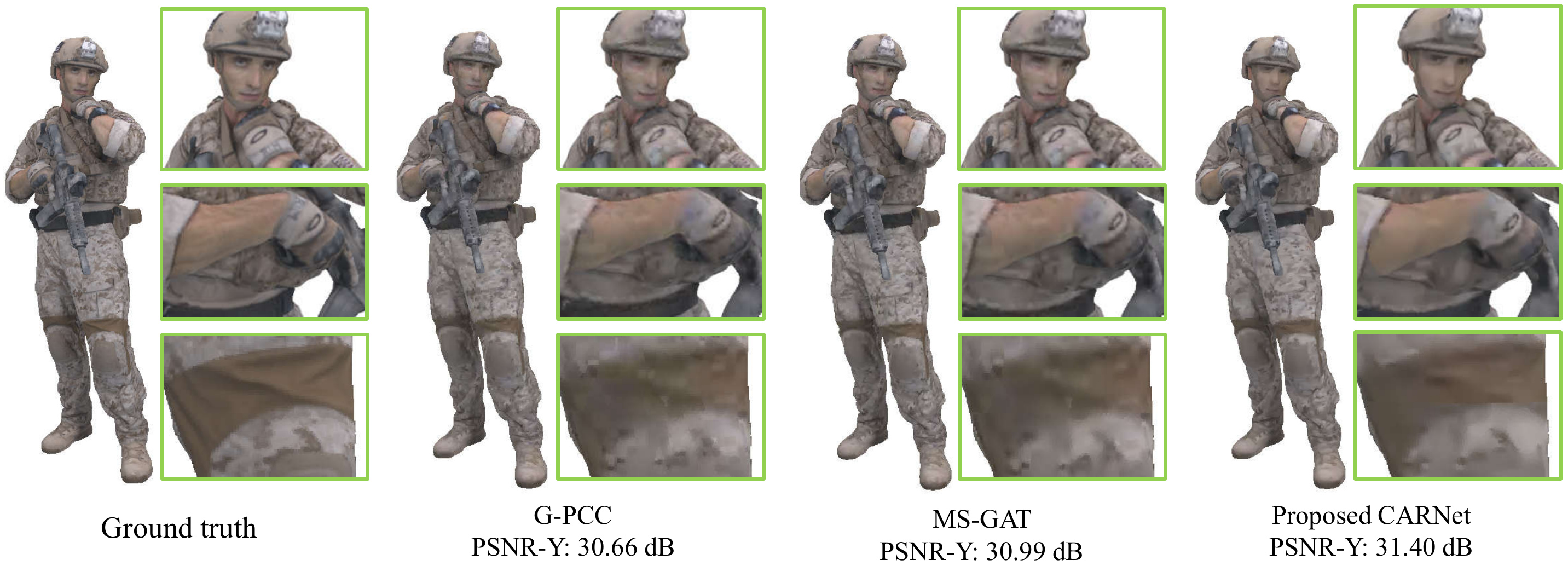} \\ \vspace{1mm}
    \includegraphics[width=0.96\linewidth]{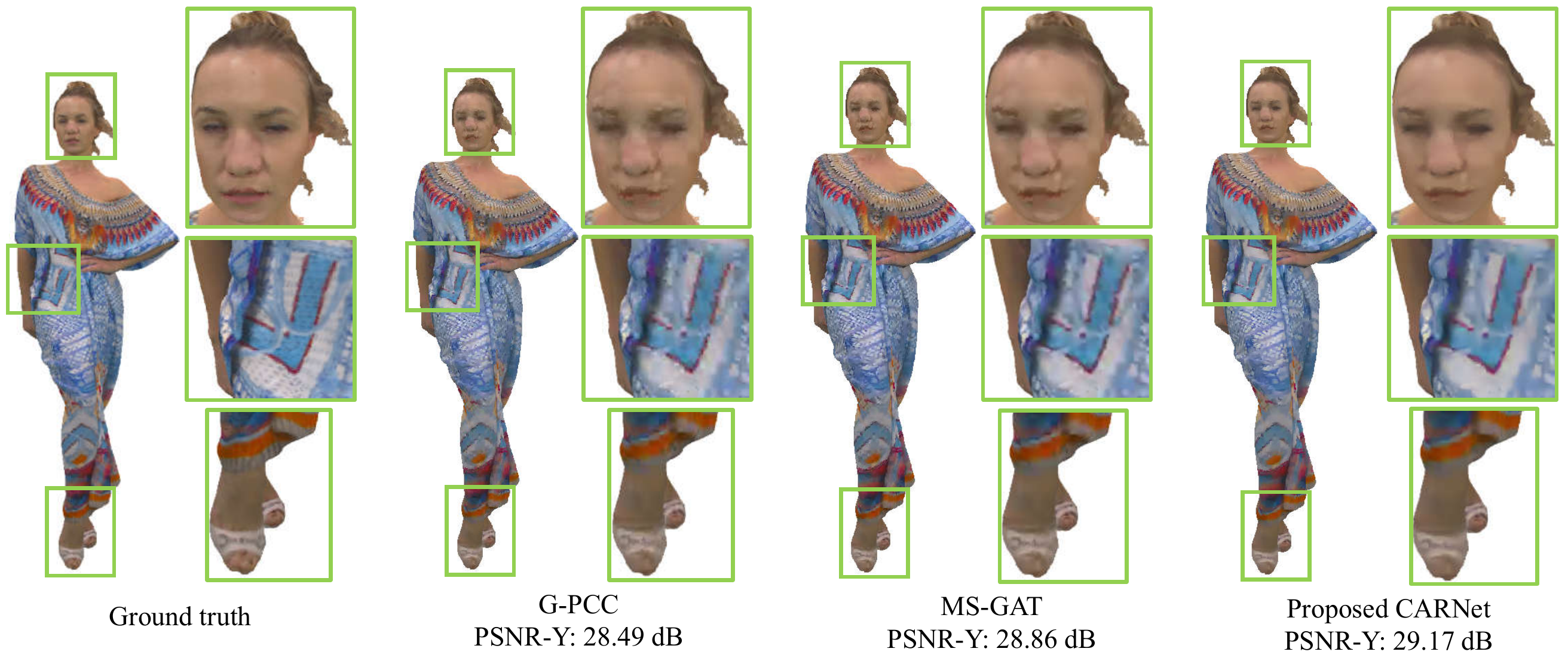} \\ \vspace{1mm}
    \includegraphics[width=0.98\linewidth]{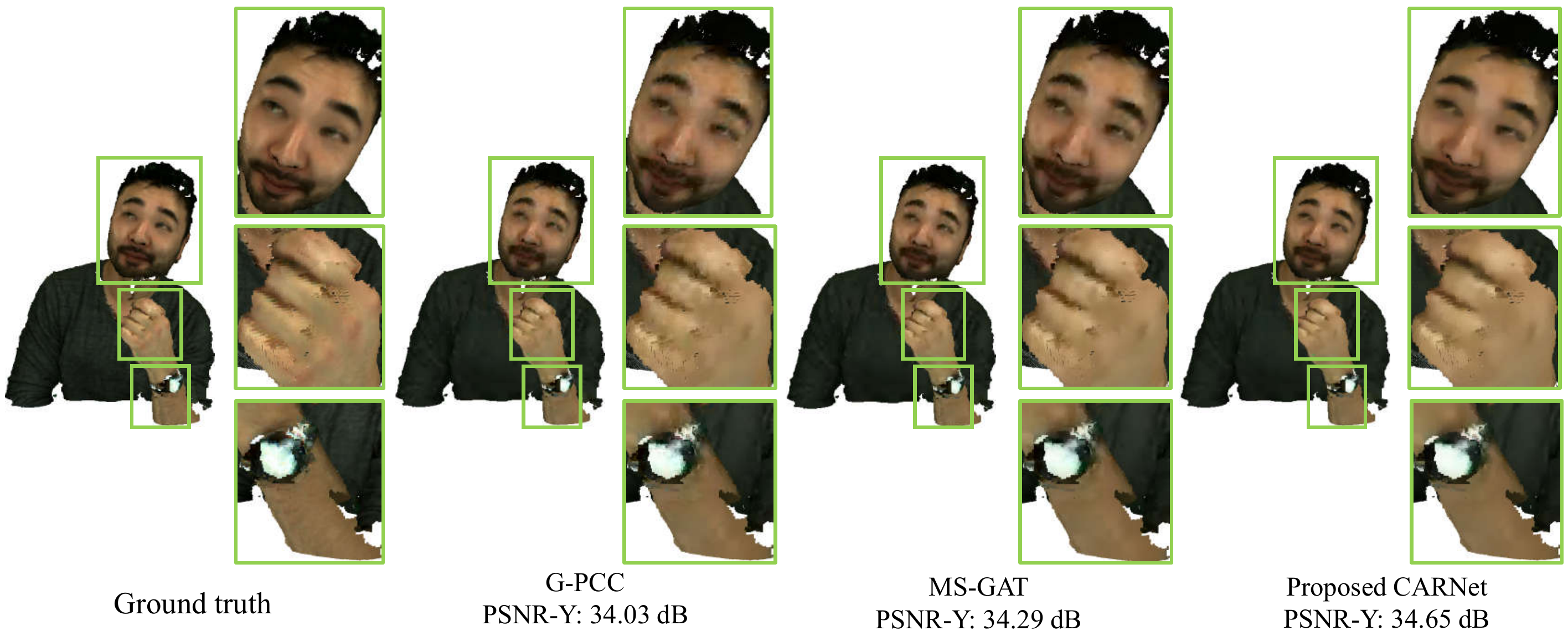}
\caption{{\bf Qualitative visualization of reconstructions using G-PCC, MS-GAT, and proposed CARNet.} The QP is 40 for G-PCC attribute compression. Three point clouds including soldier, longdress, and david are visualized from the top to bottom.} 
\label{fig:compare_visual}
\end{figure*}

We further visualize  reconstructed PCA samples that are processed using the G-PCC, the MS-GAT, and the proposed CARNet for comparison in Fig.~\ref{fig:compare_visual}. 
Three examples are compressed by G-PCC using QP value 40. The reconstructed attributes from the G-PCC contain inevitable compression noise with noticeable blurriness and blockiness, for example, the face of ``longdress'' is severely distorted and exhibits obvious blockiness. %It is the same case with the soldier's leg and foot areas. 
There is also a serious color casting issue in G-PCC reconstructions, e.g., the soldier's face. 

Enhanced by the MS-GAT, the quality of reconstructed attributes is improved to some extent, yielding more smoothing and less noisy appearance. However, blurriness and blockiness are still presented, such as the leg area of ``soldier'', the dress texture of ``longdress'', and the face of ``david''. Taking the ``longdress'' as an example, the lines and edges in the dress are blurry, especially the red line color is impaired apparently. 
By comparing the visualization results of G-PCC and MS-GAT, we find that the MS-GAT inherits considerable compression distortion from G-PCC, suggesting that these artifacts are not sufficiently removed and there is room for further improvement. 
  
By contrast, our reconstructions demonstrate drastic visual improvement over previous methods. As seen, the faces of all three examples look more natural and smooth, and the color is closer to the ground truth; the edges and lines are well preserved, such as the areas on the soldier's legs and the lady's dress. Since the proposed CARNet optimizes the attribute based on the entire point cloud, the restored attribute uniformly looks more appealing and realistic.

\begin{table*}[t]
\renewcommand{\arraystretch}{\mytablespace} 
  \centering
  \caption{Runtime (in seconds) analysis and comparison. Both the MS-GAT~\cite{sheng2022attribute} and the CARNet are run on the same GPU platform for runtime collection. Part\&Comb stands for block patch ``Partition and Combination'' in the MS-GAT. MD and MC denote the MPSOs Derivation and the MPSOs Combination. }
  \label{tab:compare_runtime}%
   %\resizebox{\linewidth}{!}
   {
    \begin{tabular}{c|c|c|c|c|c|ccc|ccc|ccc}
\hline
\multirow{3}{*}{Point Cloud} & \multirow{3}{*}{\begin{tabular}[c]{@{}c@{}}G-PCC\\ Dec.\end{tabular}} &\multicolumn{4}{c|}{MS-GAT} & \multicolumn{9}{c}{Proposed CARNet} \\ \cline{3-15} 
& & {Part\&} & \multirow{2}{*}{Y} & \multirow{2}{*}{U} & \multirow{2}{*}{V} & \multicolumn{3}{c|}{Y} & \multicolumn{3}{c|}{U} & \multicolumn{3}{c}{V} \\
&  & {Comb} &  &  &  & overall & MD & {MC} & overall & MD & {MC} & overall & MD & MC \\ \hline
MVUB 9-bit & 0.48 &  {0.15} &  {16.61} &  {16.42} &  {16.18} &  {1.12} &  {0.75} & { {0.37}} &  {1.11} &  {0.74} & { {0.37}} &  {1.17} &  {0.79} &  {0.38} \\ 
8iVFB 10-bit & 1.46 &  {1.68} &  {52.31} &  {50.61} &  {48.50} &  {1.66} &  {1.23} &  {0.43} &  {1.73} &  {1.29} & {0.44} &  {1.60} &  {1.32} &  {0.41} \\ 
Owlii 11-bit & 3.60 &  {15.38} &  {139.85} &  {137.16} &  {138.03} &  {2.80} &  {2.40} & {0.41} &  {2.92} &  {2.50} & {0.42} &  {2.99} &  {2.58} &  {0.42} \\ \hline
{\bf Average} & {\bf 1.40}  & {\bf 3.66} & {\bf 51.97} & {\bf 50.82} & {\bf 50.25} & {\bf 1.63}  & {\bf 1.22}  & {\bf 0.39}   & {\bf 1.68}  & {\bf 1.26}  & {\bf 0.40}   & {\bf 1.71}  & {\bf 1.30}  & {\bf 0.40} \\ \hline
\end{tabular}
    }
  \label{tab:addlabel}
\end{table*}

%{\bf Computational Complexity Comparison.}
\subsubsection{Space and Time Complexity}
%The number of total parameters of MS-GAT is 1.98M. 
The size of each CARNet model for enhancing the Y, U, or V component is around 2.6 MB; the MS-GAT model size is about 1.8 MB for each individual color component. 

Runtime analysis is given in Table~\ref{tab:compare_runtime}. Note that since both MS-GAT and CARNet run Python codes on GPU while the C/C++ implementation of G-PCC runs on CPU instead, the collected runtime is just served as an intuitive reference to have a general idea about the computational complexity.

As seen, in the CARNet, the averaged runtime for enhancing Y, U, and V attributes is about 1.67 seconds, which is in the same order of magnitude as that of the G-PCC compression. The processing of MPSOs derivation consumes about 80\% time due to the use of stacked DNNs. The MS-GAT costs averaged 51.97, 50.82, and 50.25 seconds for Y, U, and V processing, respectively, on ten testing point clouds. Additionally, it requires 3.66 seconds for partition and combination to enforce block patch-based processing. Apparently, our CARNet speedup the MS-GAT by about 30$\times$ for the current implementation. In the future, block parallelism may be applied to accelerate the MS-GAT for faster computation.

\subsubsection{Discussion}
As seen, the proposed CARNet effectively alleviates attribute artifacts in G-PCC compressed PCAs. It not only shows performance improvement to the G-PCC anchor but also presents a large margin to state-of-the-art MS-GAT. We believe the performance gains introduced by the CARNet mostly come from the effective modeling of spatial variations in a local neighborhood to derive potential distortion approximations, and the use of weighting coefficient to combine these approximations for optimal compensation. On the other hand, since global graph attention is used in the MS-GAT, it has to slice the large-scale point clouds into smaller patches for processing due to unaffordable memory consumption, which in return easily falls into the local optimization trap. Instead, the CARNet can process the entire large-scale point cloud for global optimization.

Moreover, the proposed CARNet attains about 30$\times$ speedup to process a compressed PCA in comparison to the MS-GAT. The high computational complexity of the MS-GAT is probably caused by the use of graph attention computation. Even for a patch with 2048 points, the graph attention layer still requires noticeable computational resources. 

Additionally, the MS-GAT is a post-processing method while the CARNet is an in-loop filtering solution that explicitly transmits weighting coefficients in bitstream. As a result, the CARNet can better adapt itself to dynamic point cloud characteristics through the use of different coefficients. This is verified in Table~\ref{tab:compare_MSGAT} where the CARNet consistently offers performance gains across all test point clouds.

\subsection{Ablation Study}
\label{subsect:ablation}
This section conducts a series of ablation studies to understand the contribution of each module in CARNet. We train all models using colorized ShapeNet in default. The comparison anchor is the latest G-PCC.

{\bf Two-Stream Network for MPSOs Derivation}.
We study the modular contribution of the two-stream network used for MPSOs derivation and results are reported in Table~\ref{tab:compare_each_stream}. We consecutively disable the Frequency-Dependent Embedding (w/o FDE) stream and Direct Spatial Embedding stream (w/o DSE) to measure the performance loss.  As seen, in comparison to the default CARNet, the performance loss is presented when disabling different modular components. For example, $>$4.2\% and $>$3.2\% BD-Rate increases are captured for ``w/o FDE'' and ``w/o DSE'', clearly showing the effectiveness of these two modules. More loss is observed by disabling the Frequency-Dependent Embedding (i.e., w/o FDE), suggesting that it is critical to model distortion artifacts across regions with different spatial frequencies.

\begin{table}[t]
\renewcommand{\arraystretch}{\mytablespace} 
  \centering
  \caption{BD-Rate contribution of modular components in the two-stream network used for MPSOs derivation. Y BD-Rate is exemplified.}
  \label{tab:compare_each_stream}%
   %\resizebox{\linewidth}{!}
   {
    \begin{tabular}{c|c|c|c|c} \hline
    {}& {Point Cloud} & {w/o FDE} & {w/o DSE}  & {Proposed} \\ 
    \hline
    \multirow{5}{*}{\begin{tabular}[c]{@{}c@{}}MVUB\\ 9-bit\end{tabular}}
    & andrew & -11.14\%  & -14.57\%     & -15.30\%  \\
    & david & -14.91\%  & -14.19\%     & -18.35\%  \\
    & phil & -16.65\%  & -18.88\%    & -21.52\%  \\
    & ricardo & -13.76\%  & -16.42\%      & -19.52\%  \\
    & sarah & -17.53\%  & -18.12\%     & -22.10\%  \\ \hline
    \multirow{5}{*}{\begin{tabular}[c]{@{}c@{}}8iVFB\\ 10-bit\end{tabular}}
    & longdress & -12.82\%  & -12.82\%    & -15.94\%  \\
    & loot &-11.76\% &-14.07\%  &-16.63\% \\
    & redandblack & -10.05\%  & -11.13\%   & -13.74\%  \\
    & queen &-21.98\% &-19.50\%  &-23.50\% \\
    & soldier & -15.44\%  & -14.97\%   & -17.13\%  \\ \hline
    \multirow{4}{*}{\begin{tabular}[c]{@{}c@{}}Owlii\\ 11-bit\end{tabular}}
    & basketball\_player &-14.47\% &-16.52\%  &-21.62\% \\
    & exercise &-14.07\% &-16.24\%  &-20.41\% \\
    & dancer & -16.14\%  & -17.16\%    & -20.87\%  \\
    & model & -13.80\%  & -14.30\%   & -17.54\%  \\ \hline
    & {\bf Average} & {\bf -14.60\%}  & {\bf -15.63\%}  & {\bf -18.87\%}  \\ \hline
    \end{tabular}%
}
\end{table}%

\begin{table*}[t]
\renewcommand{\arraystretch}{1.1} 
  \centering
  \caption{Efficiency of the MPSOs Combination. Y BD-Rate is measured for analysis. Bit rate overhead for compressing weighting coefficients is measured in bpp.}
  \label{tab:compare_guided}%
   \resizebox{\linewidth}{!}
{
\begin{tabular}{c|c|c|cc|cc|cc|cc} \hline
\multirow{3}{*}{}&\multirow{3}{*}{Point Cloud} & \multirow{2}{*}{w/o MPSOs} & \multicolumn{6}{c}{w/ MPSOs Combination} \\ \cline{4-11}
 & & \multirow{2}{*}{Combination} & \multicolumn{2}{c|}{1 weighting coefficient} & \multicolumn{2}{c|}{2 weighting coefficients} & \multicolumn{2}{c|}{3 weighting coefficients} & \multicolumn{2}{c}{4 weighting coefficients} \\ 
 & & & BD-Rate & bpp & BD-Rate & bpp & BD-Rate & bpp & BD-Rate & bpp \\ \hline
\multirow{5}{*}{\begin{tabular}[c]{@{}c@{}}MVUB\\ 9-bit\end{tabular}}
& andrew & -7.64\% & -14.03\% & 2.04E-05 & -13.34\% & 3.80E-05 & -15.3\% & 5.57E-05 & -14.65\%  & 7.30E-05 \\
& david & +0.56\% & -16.96\% & 1.59E-05 & -17.13\% & 3.40E-05 & -18.35\% & 4.74E-05 & -16.9\%   & 6.30E-05 \\
& phil & -5.34\% & -20.89\% & 1.51E-05 & -20.76\% & 2.80E-05 & -21.52\% & 4.27E-05 & -20.07\%  & 5.60E-05 \\
& ricardo & +1.65\% & -17.16\% & 2.69E-05 & -17.96\% & 4.90E-05 & -19.52\% & 7.30E-05 & -19.98\%  & 1.00E-04 \\
& sarah & -3.09\% & -20.28\% & 2.16E-05 & -20.12\% & 3.50E-05 & -22.1\% & 5.20E-05 & -21.73\%  & 7.46E-05 \\ \hline
\multirow{5}{*}{\begin{tabular}[c]{@{}c@{}}8iVFB\\ 10-bit\end{tabular}}
& longdress & -5.88\% & -14.74\% & 7.46E-06 & -14.82\% & 1.30E-05 & -15.94\% & 2.10E-05 & -15.44\%  & 2.70E-05 \\
& loot &-1.07\%  &-15.76\%  &6.21E-06  &-15.62\%  &1.24E-05  &-16.63\%  &1.86E-05  &-15.50\%   &2.48E-05 \\
& redandblack & -2.09\% & -13.47\% & 2.00E-06 & -13.79\% & 1.40E-05 & -13.74\% & 2.17E-05 & -14.22\%  & 2.80E-05 \\
& queen &-4.67\%  &-22.33\%  &5.00E-06  &-22.83\%  &1.01E-05  &-23.50\%  &1.50E-05  &-23.79\%   &1.99E-05 \\
& soldier & -3.93\% & -15.79\% & 5.47E-06 & -16.05\% & 1.00E-05 & -17.13\% & 1.42E-05 & -16.68\%  & 2.00E-05 \\ \hline
\multirow{4}{*}{\begin{tabular}[c]{@{}c@{}}Owlii\\ 11-bit\end{tabular}}
& basketball\_player & +1.00\%  &-19.99\%  &1.71E-06  &-20.28\%  &3.42E-06  &-21.62\%  &5.13E-06  &-21.09\%   &6.84E-06 \\
& exercise & +2.87\%  &-18.86\%  &2.09E-06  &-19.02\%  &4.18E-06  &-20.41\%  &6.27E-06  &-20.09\%   &8.36E-06 \\
& dancer & -0.99\% & -21.46\% & 2.00E-06 & -20.27\% & 4.00E-06 & -20.87\% & 6.00E-06 & -21.46\%  & 8.00E-06 \\
& model & -1.88\% & -17.82\% & 2.07E-06 & -17.21\% & 4.08E-06 & -17.54\% & 6.12E-06 & -17.82\%  & 8.16E-06 \\ \hline
& {\bf Average} & {\bf -2.18\%} & {\bf -17.83\%} & {\bf 1.47E-05} & {\bf -17.76\%} & {\bf 1.85E-05} & {\bf -18.87\%} & {\bf 2.75E-05} & {\bf -18.49\%} & {\bf 3.70E-05} \\ \hline 
\end{tabular}
}
\end{table*}

{\bf MPSOs Combination Mechanism.}
The MPSOs Combination using linear weighting is first examined by using a conventional single-channel output. As such, no weighting coefficient is encoded and transmitted. As shown in Table~\ref{tab:compare_guided}, the option of ``w/o MPSOs Combination'' gains only 2.18\% on the Y component, which is much lower than the default CARNet with MPSOs Combination (see Table~\ref{tab:compare_GPCC}).  More importantly, BD-rate increases are shown for point clouds like ``david'', ``ricardo'', ``basketball\_player, and ``exercise''. All of these clearly reveal that the MPSOs Combination can well capture content dynamics for better generalization.

Moreover, we also investigate the enhancement performance of the CARNet using a different number of linear weighting coefficients (or equivalently, a different number of MPSOs). For example, in Table~\ref{tab:compare_guided}, we report the results of using 2, 3, and 4 weighting coefficients. In general, using 3 and 4 coefficients have comparable gains, which is better than using 1 or 2 coefficients. In this paper, we hence adopt the use of 3 weighting coefficients. 
Besides, the bit cost of weighting coefficients is reported. It is obvious that the average bit cost of coefficients is negligible, implying almost zero impact on the overall bit rate consumption of compressed PCAs.

{\bf  Cross-Component Strategy to Train Separate CARNet Models for Chroma Components.}
%YUV and Y, U, V alone'
In Table~\ref{tab:compare_cross}, we report the results of training a joint CARNet model for Y, U, and V components. Since the Y, U, and V components possess very diverse characteristics and distribution, a joint model may compromise the performance to some extent. This is confirmed by our experiments: the BD-Rate gain of the joint model of Y, U, and V is decreased to 10.64\% in average. On the contrary, the proposed CARNet which trains Y, U, and V separately with the cross-component strategy for chroma attributes achieves 21.96\% BD-Rate reduction. 

\begin{table}[t]
\renewcommand{\arraystretch}{1.1} 
  \centering
  \caption{Joint YUV model versus Separate Y, U, V models. Cross-component strategy is applied to train the CARNet for U and V. BD-Rate is measured in YUV space.}
  \label{tab:compare_cross}%
   %\resizebox{\linewidth}{!}
{
    \begin{tabular}{c|c|c|c} \hline
    {} & Point Cloud & Joint Model  & Separate Model \\ \hline
    \multirow{5}{*}{\begin{tabular}[c]{@{}c@{}}MVUB\\ 9-bit\end{tabular}}
    & andrew & -11.32\% & -18.36\% \\
    & david & -11.96\% & -23.38\% \\
    & phil & -15.5\% & -23.16\% \\
    & ricardo & -11.71\% & -22.32\% \\
    & sarah & -15.08\% & -25.54\% \\ \hline
    \multirow{5}{*}{\begin{tabular}[c]{@{}c@{}}8iVFB\\ 10-bit\end{tabular}}
    & longdress & -9.05\% & -19.20\% \\
    & loot &-5.72\% &-20.15\% \\
    & redandblack & -8.35\% & -17.38\% \\
    & queen &-13.78\% &-31.61\% \\ 
    & soldier & -4.51\% & -19.25\% \\ \hline
    \multirow{4}{*}{\begin{tabular}[c]{@{}c@{}}Owlii\\ 11-bit\end{tabular}}
    & basketball\_player &-9.42\% &-21.19\% \\
    & exercise &-7.94\% &-21.82\% \\
    & dancer & -12.46\% & -21.74\% \\
    & model & -12.11\% & -22.27\% \\ \hline
    &{\bf Average} &{\bf -10.64\%} & {\bf -21.96\%} \\ \hline
    \end{tabular}%
}
\end{table}%

{\bf G-PCC PredLift configuration.} We further provide the performance of CARNet when using the PredLift mode for lossy G-PCC compression. We apply exactly the same method in Section~\ref{subsect:experimental_settings} to build the training dataset for deriving PredLift models. The results are shown in Table~\ref{tab:compare_PredLift}. As seen, the proposed CARNet consistently outperforms the G-PCC by 16.53\% and the MS-GAT by 6.47\% BD-Rate, further demonstrating its generalizability for different compression settings.

\begin{table}[htbp]
\renewcommand{\arraystretch}{\mytablespace} 
  \centering
  \caption{BD-Rate (Y) gain of proposed CARNet compared with the G-PCC (TMC13v14)  and MS-GAT~\cite{sheng2021deep} in PredLift mode}
  \label{tab:compare_PredLift}%
   %\resizebox{\linewidth}{!}
   {
\begin{tabular}{c|c|c|c}\hline
  &  {Point Cloud} & {G-PCC~(TMC13v14)} & {MS-GAT} \\ \hline
% &  & BD-Rate  & BD-rate & BD-psnr \\ \hline
\multirow{5}{*}{\begin{tabular}[c]{@{}c@{}}MVUB\\ 9-bit\end{tabular}} 
 & andrew  & -10.63\%  & -5.38\%  \\
 & david & -14.4\% & -6.58\%    \\
 & phil  & -23.85\% & -11.5\%     \\
 & ricardo  & -16.42\% & -8.15\%     \\
 & sarah &  -19.32\% & -7.31\%      \\ \hline
\multirow{3}{*}{\begin{tabular}[c]{@{}c@{}}8iVFB\\ 10-bit\end{tabular}} 
 & longdress  & -15.17\%  & -2.69\%    \\
 & redandblack  & -12.59\% & -4.98\%     \\
 & soldier & -14.92\%  & -3.68\%     \\ \hline
\multirow{2}{*}{\begin{tabular}[c]{@{}c@{}}Owlii\\ 11-bit\end{tabular}} 
 & dancer  & -20.56\% & -9.09\%     \\
 & model  & -17.45\% & -5.3\%     \\ \hline
  & \textbf{Average} & \textbf{-16.53\%}  & \textbf{-6.47\%}   \\ \hline
\end{tabular}
}
\end{table}

\section{Conclusion}
The Standardized G-PCC compactly represents the geometry coordinates and color attributes of underlying  point clouds with a significant reduction of data volume at the expense of visually annoying compression artifacts. This work thus develops a learning-based adaptive in-loop filter for G-PCC, termed CARNet, to effectively alleviate attribute artifacts for G-PCC compressed point clouds.  The CARNet first leverages a two-stream network to generate a group of Most-Probable Sample Offsets (MPSOs) as compression distortion approximations and then linearly combines these MPSOs to best compensate the distortion. The linear weighting coefficients, which are estimated on the fly through the least square error optimization guided by the original input attribute, ensure optimal distortion compensation. In our implementation, we leverage sparse convolution to construct the CARNet due to the superior performance of sparse tensor in representing unorganized, irregular points. Experimental results show that the proposed CARNet achieves 21.96\% and 12.95\% BD-Rate reduction against the G-PCC and state-of-the-art MS-GAT method. Ablation studies further provide evidence for the efficiency and generalization of the proposed method in various content and compression settings.

\section{Acknowledgement}
Our sincere gratitude is directed to the authors of MS-GAT~\cite{sheng2022attribute} for making the pretrained models publicly accessible, which greatly helps us to do comparative studies.

\ifCLASSOPTIONcaptionsoff
  \newpage
\fi

\bibliographystyle{IEEEtran}
\bibliography{pcc_attribute,filter}

\begin{thebibliography}{10}
\providecommand{\url}[1]{#1}
\csname url@samestyle\endcsname
\providecommand{\newblock}{\relax}
\providecommand{\bibinfo}[2]{#2}
\providecommand{\BIBentrySTDinterwordspacing}{\spaceskip=0pt\relax}
\providecommand{\BIBentryALTinterwordstretchfactor}{4}
\providecommand{\BIBentryALTinterwordspacing}{\spaceskip=\fontdimen2\font plus
\BIBentryALTinterwordstretchfactor\fontdimen3\font minus
  \fontdimen4\font\relax}
\providecommand{\BIBforeignlanguage}[2]{{%
\expandafter\ifx\csname l@#1\endcsname\relax
\typeout{** WARNING: IEEEtran.bst: No hyphenation pattern has been}%
\typeout{** loaded for the language `#1'. Using the pattern for}%
\typeout{** the default language instead.}%
\else
\language=\csname l@#1\endcsname
\fi
#2}}
\providecommand{\BIBdecl}{\relax}
\BIBdecl

\bibitem{haala2008mobile}
N.~Haala, M.~Peter, J.~Kremer, and G.~Hunter, ``Mobile {L}i{DAR} mapping for
  3{D} point cloud collection in urban areas—a performance test,'' \emph{The
  International Archives of the Photogrammetry, Remote Sensing and Spatial
  Information Sciences}, vol.~37, pp. 1119--1127, 2008.

\bibitem{schwarz2018emerging}
S.~Schwarz, M.~Preda, V.~Baroncini, M.~Budagavi, P.~Cesar, P.~A. Chou, R.~A.
  Cohen, M.~Krivoku{\'c}a, S.~Lasserre, Z.~Li \emph{et~al.}, ``Emerging {MPEG}
  standards for point cloud compression,'' \emph{IEEE Journal on Emerging and
  Selected Topics in Circuits and Systems}, vol.~9, no.~1, pp. 133--148, 2018.

\bibitem{BDrate}
G.~Bj{\o}ntegaard, ``Calculation of average {PSNR} differences between
  rd-curves,'' in \emph{ITU-T SG 16/Q6, 13th VCEG Meeting}.\hskip 1em plus
  0.5em minus 0.4em\relax document VCEG-M33, April 2001.

\bibitem{cao2021compression}
C.~Cao, M.~Preda, V.~Zakharchenko, E.~S. Jang, and T.~Zaharia, ``{C}ompression
  of sparse and dense dynamic point clouds—methods and standards,''
  \emph{Proceedings of the IEEE}, vol. 109, no.~9, pp. 1537--1558, 2021.

\bibitem{cao20193d}
C.~Cao, M.~Preda, and T.~Zaharia, ``3{D} point cloud compression: {A} survey,''
  in \emph{The 24th International Conference on 3D Web Technology}, 2019, pp.
  1--9.

\bibitem{graziosi_nakagami_kuma_zaghetto_suzuki_tabatabai_2020}
D.~Graziosi, O.~Nakagami, S.~Kuma, A.~Zaghetto, T.~Suzuki, and A.~Tabatabai,
  ``An overview of ongoing point cloud compression standardization activities:
  video-based ({V}-{PCC}) and geometry-based ({G}-{PCC}),'' \emph{APSIPA
  Transactions on Signal and Information Processing}, vol.~9, p. e13, 2020.

\bibitem{sullivan2012overview}
G.~J. Sullivan, J.-R. Ohm, W.-J. Han, and T.~Wiegand, ``Overview of the high
  efficiency video coding ({HEVC}) standard,'' \emph{IEEE Transactions on
  Circuits and Systems for Video Technology}, vol.~22, no.~12, pp. 1649--1668,
  2012.

\bibitem{VVC_overview}
B.~Bross, J.~Chen, J.-R. Ohm, G.~J. Sullivan, and Y.-K. Wang, ``Developments in
  international video coding standardization after {AVC}, with an overview of
  versatile video coding ({VVC}),'' \emph{Proceedings of the IEEE}, vol. 109,
  no.~9, pp. 1463--1493, 2021.

\bibitem{de2016compression}
R.~L. De~Queiroz and P.~A. Chou, ``Compression of 3{D} point clouds using a
  region-adaptive hierarchical transform,'' \emph{IEEE Transactions on Image
  Processing}, vol.~25, no.~8, pp. 3947--3956, 2016.

\bibitem{VVC_Inloop}
M.~Karczewicz, N.~Hu, J.~Taquet, C.-Y. Chen, K.~Misra, K.~Andersson, P.~Yin,
  T.~Lu, E.~François, and J.~Chen, ``{VVC} in-loop filters,'' \emph{IEEE
  Transactions on Circuits and Systems for Video Technology}, vol.~31, no.~10,
  pp. 3907--3925, 2021.

\bibitem{ma2020mfrnet}
D.~Ma, F.~Zhang, and D.~R. Bull, ``{MFRN}et: a new {CNN} architecture for
  post-processing and in-loop filtering,'' \emph{IEEE Journal of Selected
  Topics in Signal Processing}, vol.~15, no.~2, pp. 378--387, 2020.

\bibitem{guan2019mfqe}
Z.~Guan, Q.~Xing, M.~Xu, R.~Yang, T.~Liu, and Z.~Wang, ``{MFQE} 2.0: A new
  approach for multi-frame quality enhancement on compressed video,''
  \emph{IEEE transactions on pattern analysis and machine intelligence},
  vol.~43, no.~3, pp. 949--963, 2019.

\bibitem{ding2021neural}
D.~Ding, X.~Gao, C.~Tang, and Z.~Ma, ``Neural reference synthesis for inter
  frame coding,'' \emph{IEEE Transactions on Image Processing}, vol.~31, pp.
  773--787, 2021.

\bibitem{nasiri2021cnn}
F.~Nasiri, W.~Hamidouche, L.~Morin, N.~Dhollande, and G.~Cocherel, ``A
  {CNN}-based prediction-aware quality enhancement framework for {VVC},''
  \emph{IEEE Open Journal of Signal Processing}, vol.~2, pp. 466--483, 2021.

\bibitem{zhang2014point}
C.~Zhang, D.~Florencio, and C.~Loop, ``Point cloud attribute compression with
  graph transform,'' in \emph{IEEE International Conference on Image Processing
  (ICIP)}.\hskip 1em plus 0.5em minus 0.4em\relax IEEE, 2014, pp. 2066--2070.

\bibitem{sheng2022attribute}
X.~Sheng, L.~Li, D.~Liu, and Z.~Xiong, ``Attribute artifacts removal for
  geometry-based point cloud compression,'' \emph{IEEE Transactions on Image
  Processing}, vol.~31, pp. 3399--3413, 2022.

\bibitem{tsai2013adaptive}
C.-Y. Tsai, C.-Y. Chen, T.~Yamakage, I.~S. Chong, Y.-W. Huang, C.-M. Fu,
  T.~Itoh, T.~Watanabe, T.~Chujoh, M.~Karczewicz \emph{et~al.}, ``Adaptive loop
  filtering for video coding,'' \emph{IEEE Journal of Selected Topics in Signal
  Processing}, vol.~7, no.~6, pp. 934--945, 2013.

\bibitem{SAO}
C.~{Fu}, E.~{Alshina}, A.~{Alshin}, Y.~{Huang}, C.~{Chen}, C.~{Tsai}, C.~{Hsu},
  S.~{Lei}, J.~{Park}, and W.~{Han}, ``Sample adaptive offset in the {HEVC}
  standard,'' \emph{IEEE Transactions on Circuits and Systems for Video
  Technology}, vol.~22, no.~12, pp. 1755--1764, 2012.

\bibitem{PCGCv2}
J.~Wang, D.~Ding, Z.~Li, and Z.~Ma, ``Multiscale point cloud geometry
  compression,'' in \emph{2021 Data Compression Conference (DCC)}.\hskip 1em
  plus 0.5em minus 0.4em\relax IEEE, 2021, pp. 73--82.

\bibitem{SparsePCGCv1}
J.~Wang, D.~Ding, Z.~Li, X.~Feng, C.~Cao, and Z.~Ma, ``Sparse tensor-based
  multiscale representation for point cloud geometry compression,'' \emph{arXiv
  preprint arXiv:2111.10633}, 2021.

\bibitem{thanh2022learning}
D.~Thanh~Nguyen and A.~Kaup, ``Learning-based lossless point cloud geometry
  coding using sparse representations,'' \emph{arXiv e-prints}, pp.
  arXiv--2204, 2022.

\bibitem{952802}
V.~Goyal, ``Theoretical foundations of transform coding,'' \emph{IEEE Signal
  Processing Magazine}, vol.~18, no.~5, pp. 9--21, 2001.

\bibitem{gu20193d}
S.~Gu, J.~Hou, H.~Zeng, H.~Yuan, and K.-K. Ma, ``3{D} point cloud attribute
  compression using geometry-guided sparse representation,'' \emph{IEEE
  Transactions on Image Processing}, vol.~29, pp. 796--808, 2019.

\bibitem{de2017transform}
R.~L. De~Queiroz and P.~A. Chou, ``Transform coding for point clouds using a
  {G}aussian process model,'' \emph{IEEE Transactions on Image Processing},
  vol.~26, no.~7, pp. 3507--3517, 2017.

\bibitem{souto2020predictive}
A.~L. Souto and R.~L. de~Queiroz, ``On predictive {RAHT} for dynamic point
  cloud coding,'' in \emph{IEEE International Conference on Image Processing
  (ICIP)}.\hskip 1em plus 0.5em minus 0.4em\relax IEEE, 2020, pp. 2701--2705.

\bibitem{quach2022survey}
M.~Quach, J.~Pang, D.~Tian, G.~Valenzise, and F.~Dufaux, ``Survey on deep
  learning-based point cloud compression,'' \emph{Frontiers in Signal
  Processing}, 2022.

\bibitem{sheng2021deep}
X.~Sheng, L.~Li, D.~Liu, Z.~Xiong, Z.~Li, and F.~Wu, ``Deep-{PCAC}: {A}n
  end-to-end deep lossy compression framework for point cloud attributes,''
  \emph{IEEE Transactions on Multimedia}, vol.~24, pp. 2617--2632, 2021.

\bibitem{he2022density}
Y.~He, X.~Ren, D.~Tang, Y.~Zhang, X.~Xue, and Y.~Fu, ``Density-preserving deep
  point cloud compression,'' in \emph{Proceedings of the IEEE/CVF Conference on
  Computer Vision and Pattern Recognition}, 2022, pp. 2333--2342.

\bibitem{PCAC}
J.~Wang and Z.~Ma, ``Sparse tensor-based point cloud attribute compression,''
  in \emph{IEEE 5th International Conference on Multimedia Information
  Processing and Retrieval}, 2022.

\bibitem{fang20223dac}
G.~Fang, Q.~Hu, H.~Wang, Y.~Xu, and Y.~Guo, ``3{DAC}: {L}earning attribute
  compression for point clouds,'' in \emph{Proceedings of the IEEE/CVF
  Conference on Computer Vision and Pattern Recognition}, 2022, pp.
  14\,819--14\,828.

\bibitem{HEVC_dblk}
A.~{Norkin}, G.~{Bjontegaard}, A.~{Fuldseth}, M.~{Narroschke}, M.~{Ikeda},
  K.~{Andersson}, M.~{Zhou}, and G.~{Van der Auwera}, ``{HEVC} deblocking
  filter,'' \emph{IEEE Transactions on Circuits and Systems for Video
  Technology}, vol.~22, no.~12, pp. 1746--1754, 2012.

\bibitem{jpeg_dong2015compression}
C.~Dong, Y.~Deng, C.~C. Loy, and X.~Tang, ``Compression artifacts reduction by
  a deep convolutional network,'' in \emph{Proceedings of the IEEE
  international conference on computer vision}, 2015, pp. 576--584.

\bibitem{wang2021multi}
Z.~Wang, C.~Ma, R.-L. Liao, and Y.~Ye, ``Multi-density convolutional neural
  network for in-loop filter in video coding,'' in \emph{2021 Data Compression
  Conference (DCC)}.\hskip 1em plus 0.5em minus 0.4em\relax IEEE, 2021, pp.
  23--32.

\bibitem{lin2022nr}
K.~Lin, C.~Jia, X.~Zhang, S.~Wang, S.~Ma, and W.~Gao, ``N{R}-{CNN}:
  Nested-residual guided {CNN} in-loop filtering for video coding,'' \emph{ACM
  Transactions on Multimedia Computing, Communications, and Applications
  (TOMM)}, vol.~18, no.~4, pp. 1--22, 2022.

\bibitem{pan2020efficient}
Z.~Pan, X.~Yi, Y.~Zhang, B.~Jeon, and S.~Kwong, ``Efficient in-loop filtering
  based on enhanced deep convolutional neural networks for {HEVC},'' \emph{IEEE
  Transactions on Image Processing}, vol.~29, pp. 5352--5366, 2020.

\bibitem{jia2021residual}
W.~Jia, L.~Li, Z.~Li, X.~Zhang, and S.~Liu, ``Residual-guided in-loop filter
  using convolution neural network,'' \emph{ACM Transactions on Multimedia
  Computing, Communications, and Applications (TOMM)}, vol.~17, no.~4, pp.
  1--19, 2021.

\bibitem{zhang2018residual}
Y.~Zhang, T.~Shen, X.~Ji, Y.~Zhang, R.~Xiong, and Q.~Dai, ``Residual highway
  convolutional neural networks for in-loop filtering in {HEVC},'' \emph{IEEE
  Transactions on Image Processing}, vol.~27, no.~8, pp. 3827--3841, 2018.

\bibitem{wang2021combining}
D.~Wang, S.~Xia, W.~Yang, and J.~Liu, ``Combining progressive rethinking and
  collaborative learning: a deep framework for in-loop filtering,'' \emph{IEEE
  Transactions on Image Processing}, vol.~30, pp. 4198--4211, 2021.

\bibitem{kong2020guided}
L.~Kong, D.~Ding, F.~Liu, D.~Mukherjee, U.~Joshi, and Y.~Chen, ``Guided cnn
  restoration with explicitly signaled linear combination,'' in \emph{2020 IEEE
  International Conference on Image Processing (ICIP)}.\hskip 1em plus 0.5em
  minus 0.4em\relax IEEE, 2020, pp. 3379--3383.

\bibitem{Cross-Component}
W.-S. Kim, W.~Pu, A.~Khairat, M.~Siekmann, J.~Sole, J.~Chen, M.~Karczewicz,
  T.~Nguyen, and D.~Marpe, ``Cross-component prediction in {HEVC},'' \emph{IEEE
  Transactions on Circuits and Systems for Video Technology}, vol.~30, no.~6,
  pp. 1699--1708, 2020.

\bibitem{he2016deep}
K.~He, X.~Zhang, S.~Ren, and J.~Sun, ``Deep residual learning for image
  recognition,'' in \emph{Proceedings of the IEEE conference on Computer Vision
  and Pattern Recognition}, 2016, pp. 770--778.

\bibitem{zhang2018enhanced}
K.~Zhang, J.~Chen, L.~Zhang, X.~Li, and M.~Karczewicz, ``Enhanced
  cross-component linear model for chroma intra-prediction in video coding,''
  \emph{IEEE Transactions on Image Processing}, vol.~27, no.~8, pp. 3983--3997,
  2018.

\bibitem{chang2015shapenet}
A.~X. Chang, T.~Funkhouser, L.~Guibas, P.~Hanrahan, Q.~Huang, Z.~Li,
  S.~Savarese, M.~Savva, S.~Song, H.~Su \emph{et~al.}, ``Shape{N}et: {A}n
  information-rich 3{D} model repository,'' \emph{arXiv preprint
  arXiv:1512.03012}, 2015.

\bibitem{lin2014microsoft}
T.-Y. Lin, M.~Maire, S.~Belongie, J.~Hays, P.~Perona, D.~Ramanan,
  P.~Doll{\'a}r, and C.~L. Zitnick, ``{M}icrosoft {COCO}: Common objects in
  context,'' in \emph{European conference on computer vision}.\hskip 1em plus
  0.5em minus 0.4em\relax Springer, 2014, pp. 740--755.

\bibitem{choy20194d}
C.~Choy, J.~Gwak, and S.~Savarese, ``4{D} spatio-temporal convnets: {M}inkowski
  convolutional neural networks,'' in \emph{Proceedings of the IEEE/CVF
  Conference on Computer Vision and Pattern Recognition}, 2019, pp. 3075--3084.

\end{thebibliography}

\end{document}